%% file: main.tex
\pdfoutput=1

\documentclass[10pt,twocolumn,letterpaper]{article}

\input{headers/header-lqa}

\input{headers/header}

\input{headers/lqa/references}

\iccvfinalcopy 

\def\iccvPaperID{8042} 

\title{Robustness Certification for Point Cloud Models}

\begin{document}


    \title{Robustness Certification for Point Cloud Models}

    \renewcommand\Authsep{,\quad}
    \renewcommand\Authand{,\quad}
    \renewcommand\Authands{,\quad}
    \renewcommand\Affilfont{\normalsize}

    \author[1]{Tobias Lorenz\thanks{This work was done while the author was at ETH Zurich.}}
    \author[2]{Anian Ruoss}
    \author[2]{Mislav Balunovi\'{c}}
    \author[3]{Gagandeep Singh}
    \author[2]{Martin Vechev}
    \affil[1]{CISPA Helmholtz Center for Information Security}
    \affil[2]{Department of Computer Science, ETH Zurich}
    \affil[3]{University of Illinois at Urbana-Champaign and VMware Research}
    \affil[ ]{\tiny}
    \affil[ ]{\tt\small tobias.lorenz@cispa.de, ggnds@illinois.edu}
    \affil[ ]{\tt\small \{anian.ruoss, mislav.balunovic, martin.vechev\}@inf.ethz.ch}

    \maketitle


    \input{abstract}
    \input{sections/introduction}
    \input{sections/background}
    \input{sections/3dcertify}
    \input{sections/deepg}
    \input{sections/taylor}
    \input{sections/maxpool}
    \input{sections/experiments}
    \input{sections/conclusion}


    \message{^^JLASTBODYPAGE \thepage^^J}


    \clearpage
    {\small
    \bibliographystyle{ieee_fullname}
    \bibliography{references}
    }


    \message{^^JLASTREFERENCESPAGE \thepage^^J}


    \ifbool{includeappendix}{%
        \clearpage
        \appendix
        \input{appendix/3d_transformations}
        \input{appendix/proof_composition}

        \input{appendix/pointnet_architectures}
        \input{appendix/additional_experiments}
        \input{appendix/maxpool_analysis}
    }{}


    \message{^^JLASTPAGE \thepage^^J}

\end{document}

%% file: headers/header-lqa.tex
\usepackage[utf8]{inputenc} 
\usepackage{lipsum} 

\usepackage[T1]{fontenc}

\usepackage{etoolbox}
\newbool{includeappendix}
\setbool{includeappendix}{true}

\input{headers/lqa/overfull}


\input{headers/lqa/acronyms}

\input{headers/lqa/listing}


%% file: headers/lqa/overfull.tex
\ifdefined\isoverfull
\else
\fi

%% file: headers/lqa/acronyms.tex
\usepackage{acro} 

\DeclareAcronym{cli} {
    short = CLI,
    long = Command Line Interface,
    class = abbrev
}

%% file: headers/lqa/listing.tex
\usepackage{listings}

\usepackage{textcomp}

\usepackage{xcolor}

\usepackage[scaled=0.8]{beramono}

\definecolor{ckeyword}{HTML}{7F0055}
\definecolor{ccomment}{HTML}{3F7F5F}
\definecolor{cstring}{HTML}{2A0099}

\lstdefinestyle{numbers}{
	numbers=left,
	framexleftmargin=20pt,
	numberstyle=\tiny,
	firstnumber=auto,
	numbersep=1em,
	xleftmargin=2em
}

\lstdefinestyle{layout}{
	frame=none,
	captionpos=b,
}

\lstdefinestyle{comment-style}{
	morecomment=[l]//,
	morecomment=[s]{/*}{*/},
	commentstyle={\color{ccomment}\itshape},
}

\lstdefinestyle{string-style}{
	morestring=[b]",%
	morestring=[b]',%
	stringstyle={\color{cstring}},
	showstringspaces=false,%
}

\lstdefinestyle{keyword-style}{
	keywordstyle={\ttfamily\bfseries},
	morekeywords={
		function,
		constructor,
		int,
		bool,
		return,
		returns,
		uint
	},
	morekeywords = [2]{},
	keywordstyle = [2]{\text},
	sensitive=true,
}

\lstdefinestyle{input-encoding}{
	inputencoding=utf8,
	extendedchars=true,
	literate=
	{ℝ}{$\reals$}1%
	{→}{$\rightarrow$}1%
	{α}{$\alpha$}1%
	{β}{$\beta$}1%
	{λ}{$\lambda$}1%
	{θ}{$\theta$}1%
	{ϕ}{$\phi$}1%
}

\lstdefinestyle{escaping}{
	moredelim={**[is][\color{blue}]{\%}{\%}},
	escapechar=|,
	mathescape=true
}

\lstdefinestyle{default-style}{
	basicstyle=\fontencoding{T1}\ttfamily\footnotesize,
	style=numbers,
	style=layout,
	style=comment-style,
	style=string-style,
	style=keyword-style,
	style=input-encoding,
	style=escaping,
	tabsize=2,
	upquote=true
}

\lstdefinelanguage{BASIC}{
	language=C++,
	style=default-style
}[keywords,comments,strings]%

%% file: headers/header.tex
\usepackage{iccv}

\usepackage{epsfig}
\usepackage{graphicx}
\usepackage{booktabs}
\usepackage{tabularx}
\usepackage{multirow}
\usepackage[caption=false]{subfig}
\usepackage{authblk}

\usepackage{amsmath}
\DeclareMathOperator*{\argmax}{arg\,max}

\usepackage{amssymb}
\usepackage{bm}
\usepackage{gensymb}

\usepackage{times}

\makeatletter
\@namedef{ver@everyshi.sty}{}
\makeatother
\usepackage{tikz}
\usepackage[pagebackref=true,breaklinks=true,letterpaper=true,colorlinks,bookmarks=false]{hyperref}

%% file: headers/lqa/references.tex
\usepackage[capitalize]{cleveref}

\crefname{listing}{Lst.}{listings}
\crefname{line}{Lin.}{Lin.}
\crefname{appendix}{App.}{App.}

\newcommand{\app}[1]{%
	\ifbool{includeappendix}{\cref{#1}}{the appendix}%
}
\newcommand{\App}[1]{%
	\ifbool{includeappendix}{\cref{#1}}{The appendix}%
}

%% file: abstract.tex
\begin{abstract}
The use of deep 3D point cloud models in safety-critical applications, such
as autonomous driving, dictates the need to certify the robustness of these
models to real-world transformations. This is technically challenging, as it
requires a scalable verifier tailored to point cloud models that handles a
wide range of semantic 3D transformations. In this work, we address this
challenge and introduce 3DCertify, the first verifier able to certify
the robustness of point cloud models. 3DCertify is based on two key insights:
(i) a generic relaxation based on first-order Taylor approximations,
applicable to any differentiable transformation, and (ii) a precise
relaxation for global feature pooling, which is more complex than
pointwise activations (\eg, ReLU or sigmoid) but commonly employed in
point cloud models. We demonstrate the effectiveness of 3DCertify by
performing an extensive evaluation on a wide range of 3D transformations
(\eg, rotation, twisting) for both classification and part segmentation
tasks. For example, we can certify robustness against rotations by
\textpm 60\textdegree\ for 95.7\% of point clouds, and our max pool relaxation
increases certification by up to 15.6\%.
\end{abstract}

%% file: sections/introduction.tex
\section{Introduction}
\label{lsec:introduction}

Deep learning has achieved remarkable success in tasks involving 3D objects
such as autonomous driving~\cite{chen20213d, chen2017multiview, liang2018deep}.
As such applications are typically safety-critical,
recent work has investigated methods for quantifying the robustness of these
systems via adversarial attacks, demonstrating the vulnerability of
state-of-the-art point cloud models to semantic transformations and noise-based
perturbations~\cite{liu2019extending, wen2019geometry, xiang2019generating,
yang2019adversarial, zhao2020isometry}.
Another line of work introduced defenses~\cite{yang2019adversarial,
zhang2019defensepointnet, zhou2019dupnet}, aiming to improve a models'
robustness against these attacks.
However, as demonstrated in the image recognition domain, such defenses are
usually broken by more powerful attacks~\cite{athalye2018obfuscated,
tramer2020adaptive}, resulting in an arms race between stronger defenses and
even stronger attacks.

To break this cycle, one ideally needs a \emph{proof} that a deep learning model
is robust against \emph{any} adversarial attack, under some threat model.
This proof is usually obtained by invoking a neural network verifier on a deep
learning model and a model input, where the verifier attempts to provide a
certificate that the model is robust to any transformation of this input.
While plenty of verifiers have been proposed in the image recognition
domain~\cite{boopathy2019cnncert, katz2017reluplex, singh2019beyond,
singh2019abstract, tjeng2019evaluating, wong2018provable, zhang2018efficient},
no such verifier exists for 3D point cloud models.

\paragraph{This work: certification of 3D point clouds}

In this work we propose the first verifier for 3D point cloud models, called 3DCertify. As point cloud models are too complex for exact verification (\eg, MILP~\cite{tjeng2019evaluating}), the key challenge one must address is designing scalable and precise convex relaxations capturing all point clouds that could result from transforming the original (input) point cloud. In our work, we address this challenge and introduce such relaxations for a large family of common differentiable 3D transformations, including rotation, twisting, tapering, shearing, and arbitrary compositions of these transformations.

Robustness certification is achieved by propagating our 3D relaxations through the network using existing verifiers (\eg, DeepPoly~\cite{singh2019abstract} or LiRPA~\cite{xu2020automatic}). This modular design allows our method to benefit from future advances in verification. For instance, we observe that existing verifiers incur a significant loss of precision at the max pool layer, often overlooked when designing verifiers for image classifiers, yet a critical feature of 3D point cloud models~\cite{qi2017pointnet}. We address this issue by designing a more precise max pool relaxation which is more precise than existing solutions in practice while being applicable beyond point cloud models.

Using 3DCertify, we are able to certify, for the first time, the robustness of the PointNet~\cite{qi2017pointnet} model to semantic transformations on two challenging tasks: object classification on the ModelNet40~\cite{wu20153d} dataset and part segmentation on the ShapeNet~\cite{chang2015shapenet} dataset. We consider PointNet since, despite its relatively simple architecture, it performs well in terms of classification, segmentation, and certification, and provides the basis for more complex models~\cite{qi2017pointnet++, wang2019dynamic}, which could be certified with future advances in verification.

\paragraph{Key contributions} Our main contributions are:

\begin{itemize}
    \item A novel framework based on first-order Taylor approximations for fast computation of linear relaxations of semantic 3D transformations of point clouds.
    \item A scalable linear relaxation for max pool that is provably more precise than prior work while being widely applicable to certification tasks beyond point clouds.
    \item The identification of inherent accuracy-robustness trade-offs in point cloud networks and a detailed ablation study of robustness-enhancing methods.
    \item The first robustness verifier of 3D point cloud models for object classification and part segmentation.
    \item A comprehensive experimental evaluation of our method on the PointNet~\cite{qi2017pointnet} architecture and different datasets using a
    variety of semantic transformations. We make our implementation publicly available at \small{\url{https://github.com/eth-sri/3dcertify}}.
\end{itemize}

%% file: sections/background.tex
\section{Background \& Related Work}
\label{seq:background-related-work}

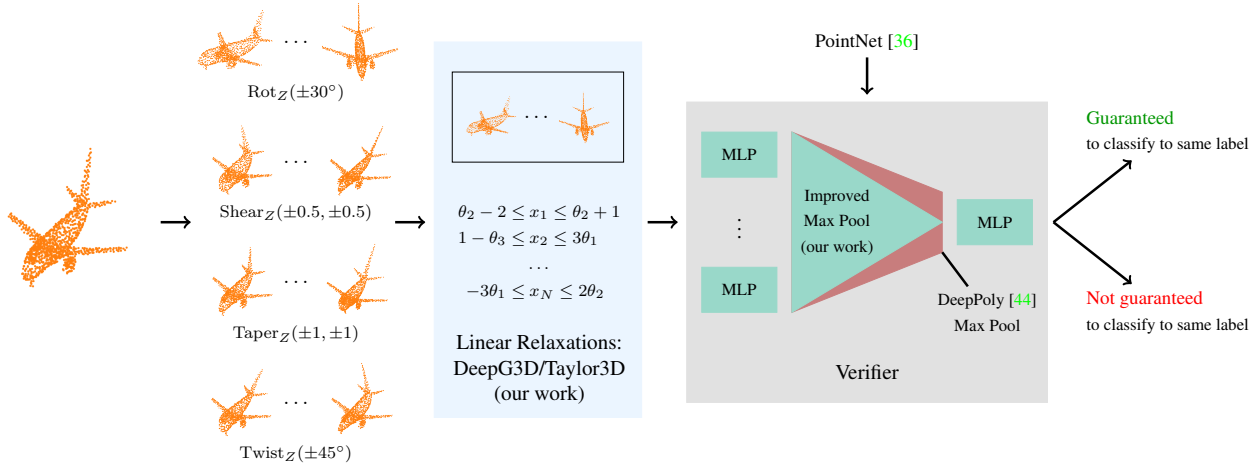
\begin{figure*}
    \begin{center}
        \resizebox{\textwidth}{!}{ \input{figures/overview.tex} }
    \end{center}
    \vspace{-4.5mm}
    \caption{
        Overview of 3DCertify, consisting of two components: (i) a method to
        compute linear relaxations of 3D transformations for point clouds, and
        (ii) an improved network verifier.
        The first stage receives a point cloud and a transformation as input and
        computes a linear relaxation around all possible transformed point clouds.
        This relaxation is passed to the second stage, where the verifier tries
        to prove that a given network correctly classifies the input object
        under all (infinitely many) transformations.
        The verifier leverages our improved relaxation for the global max pool
        commonly employed in 3D models, enabling certification of significantly
        more objects than prior work.
    }
    \label{fig:overview}
	\vspace{-4.5mm}
\end{figure*}

We now provide the necessary background on 3D point clouds and neural network
certification, and give an overview of work closely related to ours in these
areas.

\paragraph{3D point cloud models}

3D data can be represented in several different ways (\eg, 2D projections, 3D
voxels, or 3D meshes), each with distinct advantages for specific applications.
In this work, we consider 3D point clouds, which represent an object as an
unordered set of points.
Point clouds are sparse, requiring less space than, \eg, voxels, and are the
natural representation of 3D sensors such as LIDAR devices.
The key challenge for neural networks handling point clouds is to be
invariant to permutations in the order of 3D points.
Qi \etal~\cite{qi2017pointnet} were the first to address this challenge with
their novel PointNet architecture, which aggregates local, pointwise features
with a symmetric function (usually max pool) to obtain a
permutation-invariant global feature vector that can be processed by any
task-specific network, including classification and part segmentation models.
In an effort to boost model performance, the PointNet architecture has been
extended to include local information~\cite{qi2017pointnet++, wang2019dynamic},
or rotation-invariant features~\cite{li2020rotation}.

\paragraph{Adversarial attacks on point cloud models}

Qi \etal~\cite{qi2017pointnet} performed an initial investigation of the
robustness of the PointNet architecture by introducing the concepts of critical
points and upper bound shapes.
They demonstrated that all sampled point clouds lying between critical points
and upper bound shapes generate the same global feature vectors and thus obtain
the same classification.
However, this form of robustness quantification only holds for the concrete
samples and does not consider adversarial transformations.
The latter problem was addressed by a recent line of work that extended the
well-studied problem of adversarial attacks for images to the 3D point cloud
domain by considering adversarial point perturbation and
generation~\cite{lang2020geometric, lee2020shapeadv, liu2019extending,
liu2020adversarial, wen2019geometry, xiang2019generating, yang2019adversarial},
real-world adversarial objects for LIDAR sensors~\cite{cao2019adversarial},
occlusion attacks~\cite{wicker2019robustness}, and adversarial
rotations~\cite{zhao2020isometry}.
The adversarial vulnerability of 3D point cloud models has spurred the
development of corresponding defense methods, based on perturbation
measurement~\cite{yang2019adversarial}, outlier removal and
upsampling~\cite{zhou2019dupnet}, and adversarial
training~\cite{liu2019extending, zhang2019defensepointnet}.
However, similar to the image setting, many of these adversarial defenses
were later broken by stronger attacks~\cite{sun2020adversarial}, illustrating
the need for provable robustness guarantees of 3D point cloud models.

\paragraph{Neural network certification}

Existing certifiers prove the robustness of image and NLP models: they compute a
certificate which guarantees that no attack in a given range can change the
predicted label (local robustness).
These approaches generally focus on $\ell_p$-norm threat models (\ie, changing
pixel intensities) and are based on the following three-step process: (i)
compute a convex shape that encloses the space of all inputs obtainable from a
given threat model, implying that the shape is \emph{sound}, (ii) propagate the
input shape through the network to obtain an output shape on the logits, and
(iii) check that all concrete outputs within that output shape are classified to
the correct class.
For smaller networks, certification can be performed exactly with SMT
solvers~\cite{katz2017reluplex} or mixed-integer linear
programming~\cite{tjeng2019evaluating}, but scaling to bigger networks requires
computing an overapproximation of the output shape which can cause false
positives, \ie, the verifier fails to prove robustness even though it holds. A
variety of such methods have been proposed based on semi-definite
relaxations~\cite{raghunathan2018semidefinite}, linear
relaxations~\cite{bonaert2021fast, boopathy2019cnncert, gehr2018ai2,
lin2019robustness, salman2019convex, singh2018fast, singh2019abstract,
wang2018formal, weng2018towards, wong2018provable, xu2020automatic,
zhang2018efficient}, or combinations of solvers and the
above~\cite{muller2021precise, singh2019beyond, singh2019boosting,
wang2018efficient}.
These overapproximation methods compute convex relaxations for nonlinear network
operations (\eg, ReLU or max pool). The main challenge in the design of these
relaxations is balancing their cost and precision.
Certification can also be performed via randomized
smoothing~\cite{rosenfeld2019certified, lecuyer2019certified,
salman2019provably}, which, however, provides probabilistic
guarantees for a smoothed version of the original model and also incurs overhead
during inference (as it requires additional sampling).

\paragraph{Certification of semantic transformations}

Beside $\ell_p$-norm threat models, neural network certification against
semantic image transformations (\eg, rotation or translation) have been recently considered.
These approaches use enumeration~\cite{pei2017towards},
intervals~\cite{singh2019abstract}, linear
relaxations~\cite{balunovic2019certifying, mohapatra2020towards,
ruoss2020efficient}, or randomized smoothing~\cite{fischer2020certification,
li2020provable}.
Inspired by randomized smoothing, concurrent work~\cite{liu2021pointguard}
computes probabilistic certificates for point cloud models against modifying a
few individual points (addition or deletion), but cannot handle the semantic
transformations considered in our work (\eg, rotations or shearing).
Another concurrent work~\cite{fischer2021scalable} introduces an alternative
approach to certify segmentation based on randomized smoothing.
To the best of our knowledge, no prior certification method can handle the
type of transformations on point cloud models which we consider.
To achieve our goal, we leverage the high-level idea of DeepG~\cite{balunovic2019certifying},
\ie, we compute linear bounds on the transformed point positions in terms of
the transformation parameters (\eg, rotation angle). DeepG computes these
bounds via a combination of sampling and optimization, which is asymptotically
optimal but can unfortunately require exponential time for 3D transformations
with many variables. Accordingly, this severely affects the practicality of the method,
as we demonstrate experimentally. Thus, in our work, we first show how to
generalize DeepG to the 3D point cloud setting and then introduce a more
efficient (constant time) relaxation framework specifically tailored to point
clouds. The asymptotic benefit is also reflected in practice, where we
demonstrate 1000x speed-ups over the DeepG generalization, with minimal or no
loss of precision.

We achieve end-to-end certification by providing our linear bounds as input to
neural network verifiers, such as DeepPoly~\cite{singh2019abstract} and
LiRPA~\cite{xu2020automatic}, which work by overapproximating common functions
(\eg, ReLU) with a linear upper and lower bound for every output
in terms of its inputs.
To further improve on our results, we develop a novel max pool relaxation
compatible with these verifiers (LiRPA does not support max pool, and DeepPoly's
relaxation is imprecise as we will show experimentally).
Finally, we show that our relaxations are generally applicable and of interest
beyond the point cloud setting.

%% file: figures/overview.tex
\begin{tikzpicture}

    \definecolor{dgreen}{RGB}{44,162,95}
    \definecolor{lgreen}{RGB}{153,216,201}
    \definecolor{dblue}{RGB}{55, 135, 255}
    \definecolor{lblue}{RGB}{235, 245, 255}
    \definecolor{lgrey}{RGB}{225,225,225}
    \definecolor{dgrey}{RGB}{100,100,100}
    \definecolor{dgreen}{RGB}{0, 150, 0}
    \definecolor{lyellow}{RGB}{255, 240, 220}
    \definecolor{lred}{RGB}{200, 125, 125}

    \node at (-3.5, 0) {\includegraphics[width=3cm]{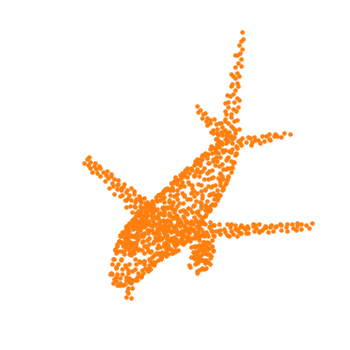}};

    \draw [->, very thick] (-1.75, 0) -- (-1.25, 0.0);

    \node at (-0.5, 3.0) {\includegraphics[width=1.5cm]{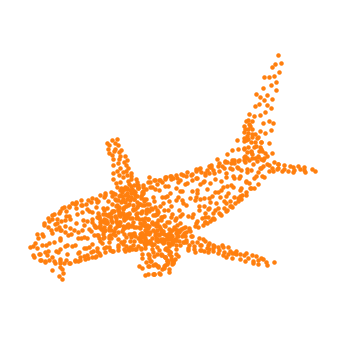}};
    \node at (0.5, 3.0) {$\ldots$};
    \node at (1.5, 3.0) {\includegraphics[width=1.5cm]{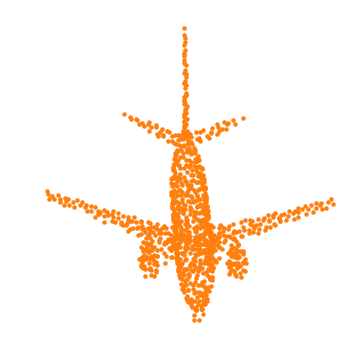}};
    \node at (0.5, 2.15) {\footnotesize $\mathrm{Rot}_Z(\pm30^\circ)$};

    \node at (-0.5, 1.0) {\includegraphics[width=1.5cm]{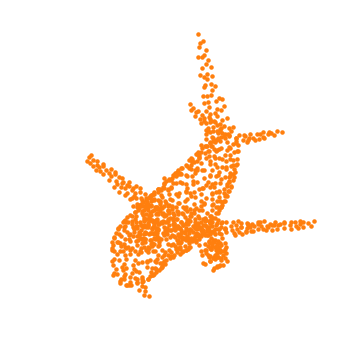}};
    \node at (0.5, 1.0) {$\ldots$};
    \node at (1.5, 1.0) {\includegraphics[width=1.5cm]{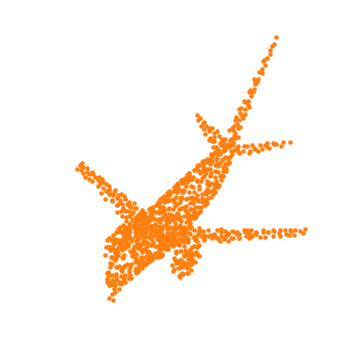}};
    \node at (0.5, 0.15) {\footnotesize $\mathrm{Shear}_Z(\pm0.5, \pm0.5)$};

    \node at (-0.5, -1.0) {\includegraphics[width=1.5cm]{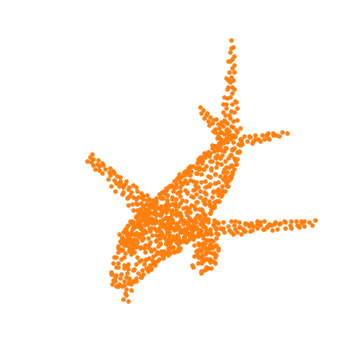}};
    \node at (0.5, -1.0) {$\ldots$};
    \node at (1.5, -1.0) {\includegraphics[width=1.5cm]{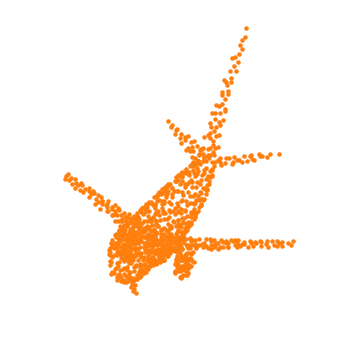}};
    \node at (0.5, -1.85) {\footnotesize $\mathrm{Taper}_Z(\pm1, \pm1)$};

    \node at (-0.5, -3.0) {\includegraphics[width=1.5cm]{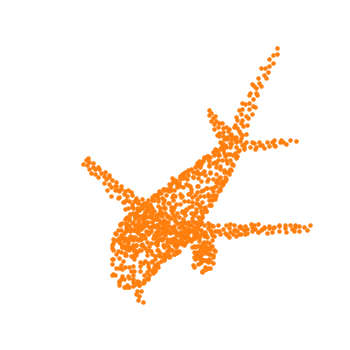}};
    \node at (0.5, -3.0) {$\ldots$};
    \node at (1.5, -3.0) {\includegraphics[width=1.5cm]{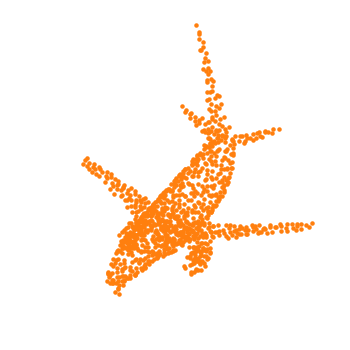}};
    \node at (0.5, -3.85) {\footnotesize $\mathrm{Twist}_Z(\pm45^\circ)$};

    \draw [->, very thick] (2.15, 0) -- (2.65, 0);

    \filldraw[fill=lblue, draw=lblue] (2.8, -3.25) rectangle (6.25, 3.0);
    \node at (3.75, 1.75) {\includegraphics[width=1cm]{figures/overview/point_cloud_24_rotated_lb}};
    \node at (4.5, 1.75) {$\ldots$};
    \node at (5.25, 1.75) {\includegraphics[width=1cm]{figures/overview/point_cloud_24_rotated_ub}};
    \node at (4.55, -0.5) {\footnotesize $\begin{aligned}
        \theta_2 - 2 \leq &~x_{1} \leq \theta_2 + 1 \\
        1-\theta_3 \leq &~x_{2} \leq 3\theta_1 \\
        &\ldots \\
        -3\theta_1 \leq &~x_{N} \leq 2\theta_2
    \end{aligned}$};
    \node[align=center] at (4.55, -2.45) {Linear Relaxations:\\DeepG3D/Taylor3D\\(our work)};

    \coordinate (e0) at (3.1, 2.5);
    \coordinate (e1) at (5.95, 2.5);
    \coordinate (e2) at (5.95, 1.0);
    \coordinate (e3) at (3.1, 1.0);
    \draw [-] (e0) -- (e1) -- (e2) -- (e3) -- cycle;

    \draw [->, very thick] (6.35, 0) -- (6.85, 0);

    \node (p) at (10.0, 3.0) {\small PointNet~\cite{qi2017pointnet}};
    \draw [->, very thick] (p) -- (10.0, 2.125);

    \filldraw[fill=lgrey, draw=lgrey] (7.0, -3.0) rectangle (13.0, 2.0);
    \node at (10.0, -2.5) {Verifier};

    \filldraw[fill=lgreen, draw=lgreen] (7.25, 0.75) rectangle (8.5, 1.5);
    \node at (7.875, 1.125) {\footnotesize MLP};
    \node at (7.875, 0.0) {$\vdots$};
    \filldraw[fill=lgreen, draw=lgreen] (7.25, -0.75) rectangle (8.5, -1.5);
    \node at (7.875, -1.125) {\footnotesize MLP};

    \coordinate (d0) at (8.75, 1.5);
    \coordinate (d1) at (8.75, -1.5);
    \coordinate (d2) at (11.25, -0.5);
    \coordinate (d3) at (11.25, 0.5);
    \filldraw[draw=lred, fill=lred] (d0) -- (d1) -- (d2) -- (d3) -- cycle;
    \node[align=center] (d) at (12.0, -1.5) {\footnotesize DeepPoly~\cite{singh2019abstract}\\\footnotesize Max Pool};
    \draw [-, very thick] (d) -- (11.25, -0.5);

    \coordinate (o0) at (8.75, 1.5);
    \coordinate (o1) at (8.75, -1.5);
    \coordinate (o2) at (11.25, 0);
    \filldraw[draw=lgreen, fill=lgreen] (o0) -- (o1) -- (o2) -- cycle;
    \node[align=center] at (9.5, 0.0) {\footnotesize Improved\\\footnotesize Max Pool\\\footnotesize (our work)};

    \filldraw[fill=lgreen, draw=lgreen] (11.5, -0.375) rectangle (12.75, 0.375);
    \node at (12.125, 0.0) {\footnotesize MLP};

    \node[align=left] (g) at (15.0, 1.5) {\small \color{dgreen} Guaranteed\\\footnotesize to classify to same label};
    \node[align=left] (n) at (15.0, -1.5) {\small \color{red} Not guaranteed\\\footnotesize to classify to same label};
    \draw [->, very thick] (13.1, 0.0) -- (g);
    \draw [->, very thick] (13.1, 0.0) -- (n);

\end{tikzpicture}

%% file: sections/3dcertify.tex
\section{3DCertify: 3D Point Cloud Verifier}
\label{sec:3dcertify}

We now present 3DCertify, our novel system for certifying the robustness of deep
learning models for point clouds against perturbations on the input data.
A high-level overview of 3DCertify is shown in \cref{fig:overview}.
Given a neural network (\eg, PointNet), a point cloud (\eg, airplane), and
a transformation (\eg, $\pm60\degree$ rotation), the goal of our system is to
prove that all transformed point clouds are classified correctly by the network.
The original point cloud and all considered transformations are shown in the
left part of \cref{fig:overview}. Verification using our system consists of two
main parts: (i) a general method for computing linear relaxations on a semantic
transformation function, and (ii) an improved network verifier to certify
the robustness of point cloud models based on these linear relaxations.
Our system is highly modular: various convex relaxations can be used in the
first part, and a wide range of network verifiers (both complete and incomplete)
can be used in the second part.

For the first part of 3DCertify, we propose two different methods for computing
linear relaxations of the transformation function, each with different benefits
and drawbacks: (a) a generalization of DeepG~\cite{balunovic2019certifying} from
images to 3D point clouds, and (b) a novel framework based on first-order
Taylor approximations.
The DeepG generalization (\cref{sec:deepg}) computes asymptotically optimal
constraints but is more expensive, especially for transformations with many
parameters.
In contrast, our Taylor relaxations (\cref{sec:taylor}) are fast to compute (in
constant time), with minimal precision loss over the constraints from DeepG.

For the second part, we leverage the state-of-the-art verifiers
DeepPoly~\cite{singh2019abstract} and LiRPA~\cite{xu2020automatic}, which scale
well to large networks in terms of computational complexity and certification
performance.
The key challenge in applying such verifiers to point cloud models is to
precisely over-approximate the large, global max pool layer.
Current 2D verifiers were designed for small pooling sizes and thus are imprecise for large pooling layers with thousands of inputs.
We address this challenge by designing a more precise max pool relaxation, which
we present in \cref{sec:maxpool}.

\paragraph{Notation}

We define a point cloud $P$ of size $n$ as
$P = \{ \bm{p^{(j)}} \mid \bm{p^{(j)}} \in \mathbb{R}^3,~j \in \{1, 2, \ldots, n\}\}$,
where each $\bm{p^{(j)}} = (x, y, z)^T$ is a point in 3D space.
We denote all vectors $\bm{v}$ in bold and their $i$-th entry as $v_i$.
A point cloud transformation is a function which changes the position of
the point cloud's points based on global transformation parameters.
Formally, we define a transformation function as
$f: \mathbb{R}^{3 \times n} \times \mathbb{R}^k \to \mathbb{R}^{3 \times n}$,
where $k$ is the number of parameters.
$f$ is applied to the point cloud prior to passing it to the neural
network.
Thus, with slight abuse of notation, given the network $N(P)$ and the
transformation function $f(P, \bm\theta)$ with parameters
$\bm\theta \in \mathbb{R}^k$, we try to certify the robustness of $N$ under
transformation $f$ for $P$ by showing that the network output
$N(f(P, \bm\theta))$ is invariant for a range of transformation parameters
$\bm{l_\theta} \leq \bm\theta \leq \bm{u_\theta}$.
To achieve this, we will compute precise linear relaxations of the
transformation function $f(P, \bm\theta)$, consisting of upper and lower
constraints $f_u$ and $f_l$.
These constraints are linear in the transformation parameters $\bm\theta$, \ie,
$f_l(P, \bm{\theta}) = \bm{\theta}^T\bm{w_l} + b_l$ and
$f_u(P, \bm{\theta}) = \bm{\theta}^T\bm{w_u} + b_u$, and must ensure that for
the parameters $\bm{l_\theta} \leq \bm\theta \leq \bm{u_\theta}$ the inequality
$f_l(P, \bm\theta) \leq f(P, \bm\theta) \leq f_u(P, \bm\theta)$ holds (both point-
and coordinatewise).

%% file: sections/deepg.tex
\subsection{Generalizing DeepG}
\label{sec:deepg}

DeepG~\cite{balunovic2019certifying} computes sound and asymptotically optimal
linear relaxations for any composition of semantic transformations, but is
limited to the 2D image domain.
Here, we show how to generalize DeepG to the 3D point cloud setting.
We only show the necessary changes, treating the remaining parts of DeepG as a
black box, and we refer interested readers to the original work for further
details.
To compute the linear relaxations, DeepG relies on a combination of sampling
and optimization, and guarantees asymptotic optimality with increasing samples
and tolerance parameter $\epsilon$.
The optimization procedure performs reverse-mode automatic differentiation and
thus requires the Jacobian of each transformation, both with respect to inputs and
parameters.
Since each point $\bm{p} \in P$ is transformed independently of the other points, we
can state the transformation in terms of $\bm{p}$ and apply it to each point
individually.
We provide the corresponding Jacobians for the 3D transformations rotation,
shearing, tapering, and twisting in \cref{app:transformations}.

%% file: sections/taylor.tex
\subsection{Taylor Approximations}
\label{sec:taylor}

In the previous section, we showed that we can compute precise linear
relaxations for point clouds by generalizing DeepG.
However, DeepG needs to solve an optimization problem for each individual
point coordinate.
While this works well for quasilinear functions with a small parameter space,
processing highly nonlinear functions with many parameters is exponentially more expensive.
We therefore propose a novel, alternative framework to compute linear constraints
in constant time based on first-order Taylor approximations.
Our method can be applied to any twice continuously differentiable transformation function
and also extends to compositions of multiple such transformations.

\paragraph{Linear approximation}

For any function $f(P, \bm\theta)$ that is differentiable on the interval
$\bm{l_\theta} \leq \bm\theta \leq \bm{u_\theta}$, the first-order Taylor
polynomial~\cite{taylor1715methodus} provides a linear approximation of $f$
around the point $\bm{t} = (\bm{u_\theta} + \bm{l_\theta}) / 2$ with approximation error
$R(P, \bm\theta)$:
\begin{equation}
	f(P, \bm\theta) = f(P, \bm{t}) + \sum_{i=1}^k{(\theta_i - t_i)\frac{\partial f}{\partial \theta_i}(P, \bm{t})} + R(P, \bm\theta).
	\label{eq:taylor}
\end{equation}
Our key insight is that we can use \cref{eq:taylor} to compute sound linear upper
and lower bounds for $f(P, \bm\theta)$, provided that we can bound the
approximation error $R(P, \bm\theta)$ with two constant terms
$L_R \leq R(P, \bm\theta) \leq U_R$, for $L_R,U_R \in \mathbb{R}^{3 \times n}$.

\paragraph{Error bounds}

For twice continuously differentiable functions, the Lagrange form of the
approximation error is
\begin{equation}
	R(P, \bm\theta) = \frac{1}{2} \sum_{i=1}^k{\sum_{j=1}^k{(\theta_i - t_i)(\theta_j - t_j)\frac{\partial^2 f}{\partial \theta_i \partial \theta_j}(P, \bm\xi)}}
\end{equation}
for some $\bm\xi$ between $\bm{t}$ and $\bm\theta$.
Using the closed intervals $\bm\theta = [\bm{l_\theta}, \bm{u_\theta}]$ and
$\bm\xi = [\bm{l_\theta}, \bm{u_\theta}]$ as inputs, we can compute the upper
and lower bounds $U_R$ and $L_R$ for $R(P, \bm\theta)$ using standard interval
arithmetic.
This means that we evaluate the functions on closed intervals instead of
concrete values, resulting in an output interval containing the entire
range of possible output values given the input intervals.
Consequently, the effect of the individual functions on the input interval is
generally overapproximated, so that the resulting interval bounds are not
necessarily exact but always sound.
Therefore, the resulting upper linear constraint
\begin{equation}
	f_u(P, \bm\theta) = f(P, \bm{t}) + \sum_{i=1}^k{(\theta_i - t_i)\frac{\partial f}{\partial \theta_i}(P, \bm{t})} + U_R
\end{equation}
and lower linear constraint
\begin{equation}
	f_l(P, \bm\theta) = f(P, \bm{t}) + \sum_{i=1}^k{(\theta_i - t_i)\frac{\partial f}{\partial \theta_i}(P, \bm{t})} + L_R
\end{equation}
create a sound overapproximation of $f$ on $[\bm{l_\theta}, \bm{u_\theta}]$.

\paragraph{Rotation example}

We demonstrate this computation with rotation around the z-axis $\mathrm{Rot}_Z$,
and we provide the relaxations for rotation around multiple axes,
as well as shearing, tapering and twisting in \cref{app:transformations}.
The transformation function for rotation around the z-axis is
\begin{equation}
	\mathrm{Rot}_Z(\bm{p}, \theta) = \begin{pmatrix}
		x \cos(\theta) - y \sin(\theta) \\
		x \sin(\theta) + y \cos(\theta) \\
		z
	\end{pmatrix}.
	\label{eq:rotation-z}
\end{equation}
Recall that each point $\bm{p} \in P$ is transformed independently, and we can
thus apply our method to each point individually.
We consider the rotation interval $\theta = [-\pi / 2, \pi / 2]$, yielding
$t = 0$ and the first-order Taylor approximation
\begin{equation}
	\mathrm{Rot}_Z(\bm{p}, \theta) = \begin{pmatrix}
		x \\
		y \\
		z
	\end{pmatrix} + \begin{pmatrix}
		-y \\
		x \\
		0
	\end{pmatrix} \theta + R(\bm{p}, \theta)
\end{equation}
with approximation error
\begin{equation}
	R(\bm{p}, \theta) = \frac{1}{2} \begin{pmatrix}
		-x \cos(\xi) + y \sin(\xi) \\
		-x \sin(\xi) - y \cos(\xi) \\
		0
	\end{pmatrix} \theta^2
\end{equation}
for $\xi \in [-\pi / 2, \pi / 2]$.
To bound this term with constant bounds $L_R \leq R(\bm{p}, \theta) \leq U_R$, we
evaluate the entire function by replacing $\xi = [-\pi / 2, \pi / 2]$ and
$\theta = [-\pi / 2, \pi / 2]$ with their valid interval range:
\begin{equation}
	[L_R, U_R] = \frac{1}{2} \begin{pmatrix}
		-x [0, 1] + y [-1, 1] \\
		-x [-1, 1] - y [0, 1] \\
		0
	\end{pmatrix} [0, \pi^2/4].
\end{equation}
Considering $\bm{p} = (1, 1, 1)^T$ as an example, we obtain
\begin{equation}
	[l_R, u_R] = \begin{pmatrix}
		[-\pi^2 / 4, \pi^2/8] \\
		[-\pi^2 / 4, \pi^2/8] \\
		0

	\end{pmatrix}.
\end{equation}

\paragraph{Composition of transformations}

Given twice continuously differentiable transformations $f(P, \bm\phi)$ and
$g(P, \bm\psi)$, the composition $h(P, \bm\theta) = g(f(P, \bm\phi), \bm\psi)$
is also twice continuously differentiable in $\bm\theta = (\bm\phi, \bm\psi)^T$
(proof in \cref{app:composition}).
Thus, we can directly compute its linear relaxation with the chain rule, using
the first- and second-order partial derivatives of $f$ and $g$.
Our framework therefore naturally extends to the composition of multiple
transformation functions.

%% file: sections/maxpool.tex
\subsection{Improved Max Pool Relaxation}
\label{sec:maxpool}

Many deep learning models for point clouds, including PointNet, use a global or
semi-global max pool layer to aggregate local, pointwise features into
permutation-invariant global features.
Consequently, these pooling layers operate on thousands of input variables,
orders of magnitude more than image classification models.
Since state-of-the-art relaxations for max pool are designed for the 2D case
with few inputs, applying these methods out-of-the-box to
point cloud models causes substantial precision loss (\cref{app:maxpool_analysis}).

Addressing the above problem by designing more precise linear constraints for
max pool is inherently difficult, as it requires reasoning in much higher
dimensions compared to most univariate activation functions (\eg, ReLU).
The state-of-the-art linear relaxations from DeepPoly for the max function
$y = \max_i x_i$ over input neurons $x_i$ with corresponding
upper and lower bounds $u_i$ and $l_i$ are: $y \geq x_j$ for $j=\argmax_i l_i$
as a lower bound and $y \leq u_{\max} = \max_i u_i$ as the upper bound.
For the lower bound, any $x_i$ would give a sound bound, since we know $y \geq x_i$.
However, we have no such guarantees for the upper bound.
DeepPoly simply uses the constant $u_{\max}$, which does not preserve relationship between neurons, causing significant precision loss.
We therefore propose a scalable and more precise upper bound which preserves
this relation.

As a first step, we check if we can prove that there exists an input neuron
$x_j$ which is always greater than the rest,
that is: $x_j > x_{i \neq j}$.
If this is the case, we simply return $y \leq x_j$ as the upper bound.
For example, consider the simplified case $y:=\max\{x_1,x_2\}$ with $x_1 \in [-1, -0.1]$ and $x_2 \in [0, 1]$.
Then $x_1 < x_2$ for all inputs and our result $y=x_2$ is exact.

Otherwise, we compute an upper bound based on the convex hull of all possible
cases $y = x_j, x_j \geq x_i$ for all possible $j$.
In the two-variate example with $x_1 \in [-1, 0.25]$ and $x_2 \in [0, 1]$,
we cannot prove $x_2 < x_1$ or $x_1 < x_2$ for all inputs and instead must
consider both cases $\mathcal{S}_1=\{y=x_1,x_1\geq x_2, x_1 \in [-1, 0.25], x_2 \in [0, 1]\}$
and $\mathcal{S}_2=\{y=x_2,x_2\geq x_1, x_1 \in [-1, 0.25], x_2 \in [0, 1]\}$.
Certifying all cases separately is prohibitive in practice, as it scales poorly
with increasing input dimension and network depth.
Instead, we compute the convex hull of $\mathcal{S}_1$ and $\mathcal{S}_2$ via the
double description method~\cite{fukuda1995double}, the state-of-the-art for
high-dimensional convex hull computation.
The double description method represents a polytope both with the set of constraints
and vertices and uses a conversion algorithms to convert between the two representations.

For vertex representations $\mathcal{V}_1$ and $\mathcal{V}_2$ of $\mathcal{S}_1$
and $\mathcal{S}_2$, the convex hull vertices are $\mathcal{V}_1 \cup \mathcal{V}_2$,
and the conversion computes output constraints
$\mathcal{S}=\{y \in [0,1], x_1 \in [-1,0.25], x_2 \in [0,1], y \geq x_1, y \geq x_2, y \leq 0.2 + 0.2 x_1 + x_2, y \leq 0.25 + 0.75 x_2\}$.
Certification with $\mathcal{S}$ is faster than considering the two cases $\mathcal{S}_1$
and $\mathcal{S}_2$ separately, but it can still be expensive due to multiple linear constraints.
Therefore, we relax $\mathcal{S}$ further by computing its DeepPoly~\cite{singh2019abstract}
relaxation, keeping the interval bounds and one upper and lower polyhedral constraint for $y$.
Prior work does not compute the convex hull and obtains $y \in [0,1]$ as interval bounds
and $x_2 \leq y \leq 1$ as polyhedral bounds.
Note that the upper polyhedral bound is the same as the upper interval bound.
In contrast, we obtain polyhedral bounds $y \leq 0.2 + 0.2 x_1 + x_2$ and
$y \leq  0.25 + 0.75 x_2$ from $\mathcal{S}$.
To choose one among them, we compute the upper bound of the right-hand-side by
replacing $x_1$ and $x_2$ with their interval bounds, obtaining 1.25 and 1, respectively.
We choose $0.25 + 0.75 x_2$ with the smaller right-hand-side as our polyhedral bound,
and our result is $y \in [0, 1]$ and $x_2 \leq y \leq 0.25 + 0.75 x_2$.
While in this example our relaxation is strictly more precise than prior work, one cannot always
theoretically establish increased precision. However, extensive experiments (\cref{sec:experiments_maxpool})
show that our method is significantly more precise in practice.

While this linear upper bound is more precise than state-of-the-art, it cannot be directly
employed to certify point cloud models, since the complexity of convex hull
computation grows exponentially in the number of input neurons, rendering the
above computation infeasible for the thousands of neurons pooled in PointNet.
Our key insight is to decompose the max pool operation into small groups of up to 10
neurons and to apply the max function recursively, making the computation tractable.
For example, to take the max over 16 neurons $\{x_1, \ldots, x_{16}\}$, we first
compute $y_1 = \max\{x_1, \ldots, x_8\}$ and $y_2 = \max\{x_9, \ldots, x_{16}\}$
and then merge both groups to obtain $y = \max\{y_1, y_2\}$.
This allows us to scale our refined relaxation to large point cloud models,
achieving significant improvements over the current state-of-the-art.
We investigate the trade-off between running time and certification precision
for different group sizes in \cref{app:max-pool}.

%% file: sections/experiments.tex
\section{Experimental Evaluation}
\label{sec:experiments}

To illustrate the broad applicability of 3DCertify, we perform extensive
experiments on different transformations, models, and tasks.
\Cref{sec:experiments_transformations} evaluates 3DCertify on a wide range
of semantic transformations, comparing our generalization of DeepG, denoted
as DeepG3D, and our Taylor framework, denoted as Taylor3D, to different
baselines and analyzing the trade-off between our two methods.
We investigate the impact of our improved max pool relaxation in
\cref{sec:experiments_maxpool}, and we conduct an ablation study of the impact of various
robustness-enhancing methods, such as adversarial training, in
\cref{sec:boosting-robustness}. Finally, we demonstrate the broad
applicability of 3DCertify by performing the first robustness certification
for part segmentation in \cref{sec:experiments_segmentation}.
We make all of our code and models publicly available at
\url{https://github.com/eth-sri/3dcertify}.

\paragraph{Experimental setup}
We use PointNet~\cite{qi2017pointnet} models for all experiments since,
despite their relatively simple architecture, they perform well for
classification, part segmentation, and certification.
Furthermore, PointNet is the basis for more complex, state-of-the-art
models~\cite{qi2017pointnet++, wang2019dynamic}, which could be certified with
future verification advancements.

For classification, we evaluate 3DCertify on the 3D objects from the
ModelNet40~\cite{wu20153d} dataset, which are represented as point clouds and
assigned to one of 40 different categories, such as tables or
airplanes.
We use the PointNet classification architecture introduced by Qi
\etal~\cite{qi2017pointnet} without T-Nets and perform experiments for
different point cloud sizes, for which we train separate versions of the model.
We apply the standard pre-processing pipeline~\cite{qi2017pointnet} to the point
clouds: centering, scaling to the unit sphere, and rotating randomly around the
z(up)-axis.
These normalizations render the model invariant to translation and scaling,
which is why we do not consider these transformations, even
though our system could easily handle them.

For certification of classification models, we report the percentage of
correctly classified objects for which we can guarantee correct classification
under all input transformations.
All reported numbers are averages over a random subset of 100 objects from the
test set (consistent with prior work~\cite{balunovic2019certifying,
mohapatra2020towards, ruoss2020efficient}).
We use the same random subset for all experiments, and, unless otherwise noted,
we run all experiments on point clouds with 64 points using our improved max
pool relaxation and the DeepPoly verifier~\cite{singh2019abstract}.

\paragraph{Splitting}

To increase certification precision for transformations with few parameters
(\eg, rotation with one angle), one can split the parameter space into
multiple smaller subsets (\eg, split rotation by $[-\theta, \theta]$ into
$[-\theta, 0]$ and $[0, \theta]$) and certify correct classification on each
subset individually.
Then, if we can certify correct classification for all subsets, we
can infer correct classification for their union.
While this technique has been successfully applied to extend certification to
larger parameter ranges~\cite{balunovic2019certifying, mohapatra2020towards,
singh2019abstract}, it can only be employed for transformations with few
parameters since the cost grows exponentially in the number of parameters.
Both DeepG3D and Taylor3D naturally support splitting.

\subsection{Semantic Transformations}
\label{sec:experiments_transformations}

Using 3DCertify, we evaluate the robustness of PointNet to several semantic
transformations: rotation around one, two, and all three axes, shearing,
twisting, and tapering, as well as compositions of these transformations.
We compare the precision of our relaxations with two baselines: simple interval
bounds, as well as refined bounds using implicit splitting introduced by
Mohapatra \etal~\cite{mohapatra2020towards} with 15\,625 splits, which we extend
to the 3D domain.
Finally, we analyze the trade-off between certification performance and
computational complexity for DeepG3D and Taylor3D.

\begin{table}
    \begin{center}
        \resizebox{\columnwidth}{!}{ \input{tables/rotation} }
    \end{center}
    \vspace{-2mm}
    \caption{
        Certification for rotation around one, two, and three axes.
    }
    \label{tbl:rotation}
    \vspace{-5mm}
\end{table}

\paragraph{Rotation}

Rotation is one of the most common transformations of 3D objects.
In particular, an object's rotation around the up-axis is often arbitrary and
thus any model should be robust to changes in upright orientation.
Additionally, a model should not be fooled by small rotations around the
other axes.
We perform thorough experiments for robustness certification of different
rotation angles around one, two, and all three axes, displaying the results in
\cref{tbl:rotation}.
All rotations are performed for $\pm \theta$, \ie, the total rotation angle is
$2\theta$, with splits of $2\degree$ along each dimension.
For rotations around a single axis, both our relaxations enable certification of
almost all objects for up to $\pm60\degree$, which is equivalent to one
third of all possible object orientations and significantly improves over
both baselines.
For two and three rotation axes, we certify robustness up to
$\pm5\degree$ and $\pm3\degree$ respectively (again outperforming both
baselines), since the number of splits required scales exponentially, as
discussed above.
Finally, to demonstrate that our system is independent of the concrete verifier,
we also instantiate the LiRPA verifier~\cite{xu2020automatic} with Taylor3D to
certify robustness against $\mathrm{Rot}_Z$ for $\pm1\degree$ without splitting.
DeepPoly certifies 97.8\% and LiRPA certifies 95.7\% of the objects, showing
that 3DCertify can effectively certify robustness of PointNet independent of
the concrete verifier used.

\paragraph{Additional transformations}
In addition to rotation, 3DCertify also handles shearing, twisting, and
tapering, as well as compositions thereof.
We compare certification accuracy and running time for DeepG3D and Taylor3D in
\cref{tbl:additional_transformations} and observe that the relaxations are
almost identical in terms of certification precision, although DeepG3D is
slightly more precise in some cases such as $\mathrm{Twist} \circ
\mathrm{Rot}_Z$.
However, Taylor3D is orders of magnitude faster, particularly for
transformations with multiple parameters.
For example, Taylor3D requires 65ms to achieve the same certification for
$\mathrm{Twist} \circ \mathrm{Taper} \circ \mathrm{Rot}_Z$ as DeepG3D, which
needs 65447ms (a 1000x speed-up).
Unlike DeepG3D, Taylor3D could thus be employed in settings requiring instant
certification feedback (\eg, real-time applications).
We further demonstrate the effortless scaling of Taylor3D to real-world point
clouds with up to 300k points \cite{geiger2013vision} in \cref{app:scaling}.

\begin{table}
    \begin{center}
        \resizebox{\columnwidth}{!}{ \input{tables/additional_transformations} }
    \end{center}
    \vspace{-2mm}
    \caption{
        Certification of different transformations and compositions thereof.
        DeepG3D provides slightly more precise certification, while Taylor3D
        is orders of magnitude more efficient.
    }
    \vspace{-1mm}
    \label{tbl:additional_transformations}
\end{table}

\begin{table}
    \begin{center}
        \resizebox{\columnwidth}{!}{ \input{tables/maxpool_rotation} }
    \end{center}
    \vspace{-2mm}
    \caption{
        Certification of $\mathrm{Rot}_Z$ with $\theta = \pm3\degree$ for
        different max pool relaxations.
        Our relaxation is significantly more precise, particularly for large
        point clouds (\ie, large pooling layers).
    }
    \label{tbl:maxpool_rotation}
    \vspace{-5mm}
\end{table}

\subsection{Improved Max Pool Relaxation}
\label{sec:experiments_maxpool}

Scaling certification to large point clouds poses significant challenges for
verifiers due to the much larger number of neurons involved in global feature
pooling, underlining the importance of a precise max pool relaxation such as the
one we introduced in \cref{sec:maxpool}.
We compare our new max pool relaxation to the DeepPoly
relaxation~\cite{singh2019abstract}, as it is the basis upon which we improve.
We also integrate the linear relaxation proposed by Boopathy
\etal~\cite{boopathy2019cnncert} into DeepPoly as an additional baseline.
We compare certification for rotation around one axis for $\pm3\degree$
without splitting on point clouds of different sizes in
\cref{tbl:maxpool_rotation}.
Our improved relaxation provides the best results in all settings, improving by
up to 15.6\% over the current state-of-the-art and with particular benefits for
large point cloud sizes, which are essential in the context of 3D point cloud
models.
We further demonstrate the broad applicability of our max pool relaxation
with experiments on computer vision models in \cref{app:max-pool}.

\subsection{Boosting Certified Robustness}
\label{sec:boosting-robustness}

It is well established that certified robustness of image
classification models can be significantly increased via adversarial and
provable training~\cite{balunovic2020adversarial, gowal2019scalable,
mirman2018differentiable, wong2018scaling, zhang2020towards}.
Here, we generalize such robust training methods to the 3D point cloud domain.
Moreover, we identify that common point cloud network components lead to
substantial trade-offs in terms of accuracy and certified robustness, providing
essential insights to guide future architecture design.

\begin{table}
    \begin{center}
        \resizebox{\columnwidth}{!}{ \input{tables/ablation} }
    \end{center}
    \vspace{-2mm}
    \caption{
        Ablation study for robustness-enhancing methods with pointwise
        $\ell_\infty$-perturbations on 64 and 256 points with
        $\epsilon = 0.01$.
    }
    \label{tbl:ablation}
    \vspace{-5mm}
\end{table}

In the 2D domain, adversarial threat models are generally formulated in terms of
$\ell_p$-norms quantifying the pixel intensity perturbation.
Such $\ell_p$-norm threat models extend naturally to the 3D point cloud setting,
where they capture the spatial perturbation of every point.
We consider the $\ell_\infty$-norm, which restricts perturbations to an
$\epsilon$-box around the original point coordinates and corresponds to the
commonly studied Hausdorff distance in the 3D adversarial attack
literature~\cite{liu2020adversarial, wen2019geometry, xiang2019generating,
yang2019adversarial}.
Note that perturbations shift each point independently from all others, unlike
\eg, rotation.

We conduct an ablation study for various robustness-enhancing methods, including
adversarial (FGSM)~\cite{goodfellow2015explaining, wong2020fast} and provable
(IBP)~\cite{gowal2019scalable} training, and using different symmetric
feature-aggregation functions (average pool and max pool).
We use the fast adversarial training variant of FGSM by Wong
\etal~\cite{wong2020fast} since it is orders of magnitude faster than
PGD~\cite{madry2018towards} (and thus applicable to point cloud models) but
provides similar performance in terms of certified robustness.
We compare the accuracy and certified robustness in \cref{tbl:ablation} and observe
that average pool leads to a notable drop in accuracy (roughly 5\%), confirming
the results by Qi \etal~\cite{qi2017pointnet}.
While naturally trained models are not provably robust against perturbations,
adversarial training significantly improves the robustness, particularly for
average pool, which DeepPoly can encode exactly regardless of the point cloud size.
In contrast, certification with max pool is challenging (particularly for larger 
point clouds)
due to its nonlinearity -- even with our improved relaxation.
Provable training further increases robustness, although less for average pool
than for max pool since IBP training uses intervals that are less precise for
average pool than for max pool.

Running the same experiment with max pool and IBP training for point clouds with
2048 points results in a model with 77.4\% accuracy and 94.9\% certification,
confirming the scalability of our approach to large point cloud sizes.

\subsection{Part Segmentation}
\label{sec:experiments_segmentation}

In addition to object classification, we consider the safety-relevant task of
part segmentation.
For part segmentation, a model tries to predict which part of an object
a point belongs to, \eg, the wings of an airplane or the legs of a chair.
We show that 3DCertify, powered by our relaxations, is the first certification
system to successfully handle this task, proving its usefulness beyond object
classification.

\begin{table}
    \begin{center}
        \resizebox{\columnwidth}{!}{ \input{tables/segmentation} }
    \end{center}
    \vspace{-2mm}
    \caption{
        Ratio of certified points for $\mathrm{Rot}_Z$ on part segmentation.
    }
    \label{tbl:segmentation}
    \vspace{-5mm}
\end{table}

For part segmentation, we use the ShapeNet-Part~\cite{chang2015shapenet}
dataset, which contains different classes of 3D objects with part annotations
for each point.
We train a part segmentation version of PointNet~\cite{qi2017pointnet} on 64
points, with an IoU score of 0.82.
The model architecture is shown in \cref{app:architectures}.
\cref{tbl:segmentation} shows the percentage of correctly classified points
remaining invariant under $\mathrm{Rot}_Z$ with $5\degree$ splits, showing that
DeepG3D and Taylor3D can certify robustness for most points, even for large
rotation angles up to $\pm10\degree$.
Moreover, the method scales to larger point clouds with 1024 points,
with 95.5\% of points certifiably robust to rotations of $\pm5\degree$.

\subsection{Discussion}

Point cloud models achieve high accuracy on natural datasets, and 3DCertify
shows that such models are also highly robust against 1D rotations, even for large
angles.
However, in line with our results, Zhao \etal~\cite{zhao2020isometry} showed
that 3D rotations achieve a 95\% attack success rate with angles as small as
$\pm2.81\degree$, implying that certification on the same model cannot scale
beyond such small angles.
In this light, our results for 3D rotations of $\pm5\degree$ are meaningful
though they highlight a need for further research on robust architectures.
Moreover, our guarantees for $\ell_\infty$-perturbations of $\epsilon=0.01$,
roughly 1\% of the object's size, indicate that 3DCertify is applicable beyond
semantic transformations.
Our robust training results demonstrate that max pool is the best symmetric
feature-aggregation function in terms of accuracy and provable robustness,
confirming its importance for point cloud models.
Finally, we note that current certifiers based on linear relaxations cannot
compute efficient bounds for PointNet's T-Nets, which are known to boost the
model's accuracy~\cite{qi2017pointnet}.
We thus consider investigating approaches to certify T-Nets without sacrificing
precision an interesting direction for future work.

%% file: tables/rotation.tex
\begin{tabular}{@{}lcrrcrrcrr@{}}
    \toprule
    && \multicolumn{2}{c}{$\mathrm{Rot}_{Z}$} && \multicolumn{2}{c}{$\mathrm{Rot}_{ZX}$} && \multicolumn{2}{c}{$\mathrm{Rot}_{ZYX}$} \\
    \cmidrule{3-4}
    \cmidrule{6-7}
    \cmidrule{9-10}
    $\pm \theta$ && 20\textdegree & 60\textdegree && 5\textdegree & 10\textdegree && 2\textdegree & 5\textdegree \\
    \midrule
    Interval && 70.7 & 54.3 && 6.5 & 3.3 && 1.1 & 0.0 \\
    Mohapatra \etal~\cite{mohapatra2020towards} && 84.8 & 80.4 && 10.9 & 4.3 && 1.1 & 1.1 \\
    Ours (DeepG3D) && \textbf{96.7} & \textbf{95.7} && \textbf{89.1} & \textbf{73.9} && \textbf{72.8} & \textbf{58.7} \\
    Ours (Taylor3D) && \textbf{96.7} & \textbf{95.7} && \textbf{89.1} & \textbf{73.9} && 69.6 & \textbf{58.7} \\
    \bottomrule
\end{tabular}

%% file: tables/additional_transformations.tex
\begin{tabular}{@{}llrr@{\extracolsep{\fill}}crr@{}}
    \toprule
    && \multicolumn{2}{c}{Certified (\%)} && \multicolumn{2}{c}{Time (ms)} \\
    \cmidrule{3-4}
    \cmidrule{6-7}
    $f(P, \theta)$ & $\pm\theta$ & Taylor3D & DeepG3D && Taylor3D & DeepG3D \\
    \midrule
    \multirow{2}{*}{$\mathrm{Rot}_{ZYX}$} & 1\textdegree & 73.9 & 73.9 && 8.87 & 391 \\
    & 1.5\textdegree & 8.7 & 8.7 && 6.88 & 405 \\
    \addlinespace
    \multirow{2}{*}{$\mathrm{Shear}$} & 0.02 & 93.5 & 93.5 && 0.07 & 230 \\
    & 0.03 & 70.7 & 70.7 && 0.07 & 228 \\
    \addlinespace
    \multirow{2}{*}{$\mathrm{Taper}$} & 0.1 & 81.5 & 81.5 && 0.46 & 232 \\
    & 0.2 & 28.3 & 28.3 && 0.47 & 232 \\
    \addlinespace
    \multirow{2}{*}{$\mathrm{Twist}$} & 10 & 76.1 & 76.1 && 0.70 & 209 \\
    & 20 & 23.9 & 23.9 && 0.64 & 211 \\
    \addlinespace
    \multirow{2}{1.3cm}{$\mathrm{Twist} \circ \mathrm{Rot}_Z$} & 10, 1\textdegree & 57.6 & 60.9 && 3.04 & 580 \\
    & 20, 1\textdegree & 7.6 & 16.3 && 3.21 & 515 \\
    \addlinespace
    \multirow{2}{1.3cm}{$\mathrm{Taper} \circ \mathrm{Rot}_Z$} & 0.1, 1\textdegree & 69.6 & 68.5 && 4.64 & 531 \\
    & 0.2, 1\textdegree & 20.7 & 20.7 && 4.54 & 918 \\
    \addlinespace
    \multirow{2}{2.1cm}{$\mathrm{Twist} \circ \mathrm{Taper} \circ \mathrm{Rot}_Z$} & 10, 0.1, 1\textdegree & 20.7 & 20.7 && 64.08 & 32216 \\
    & 20, 0.2, 1\textdegree & 5.4 & 5.4 && 65.04 & 65447 \\
    \bottomrule
\end{tabular}

%% file: tables/maxpool_rotation.tex
\begin{tabular}{@{}lcrrrrrrr@{}}
    \toprule
    Points && 16 & 32 & 64 & 128 & 256 & 512 & 1024 \\
    \midrule
    Boopathy \etal~\cite{boopathy2019cnncert} && 3.7 & 3.6 & 3.3 & 2.2 & 4.4 & 5.6 & 6.7 \\
    DeepPoly~\cite{singh2019abstract} && 95.1 & \textbf{94.0} & 91.3 & 72.2 & 51.1 & 39.3 & 28.1 \\
    Ours (Taylor3D) && \textbf{97.5} & \textbf{94.0} & \textbf{93.5} & \textbf{81.1} & \textbf{66.7} & \textbf{49.4} & \textbf{37.1} \\
    \bottomrule
\end{tabular}

%% file: tables/ablation.tex
\begin{tabular}{@{}llcrrcrr@{}}
    \toprule
    & && \multicolumn{2}{c}{Accuracy (\%)} && \multicolumn{2}{c}{Certified (\%)} \\
    \cmidrule{4-5}
    \cmidrule{7-8}
    Global Pooling & Training & & 64 & 256 && 64 & 256 \\
    \midrule
    \multirow{3}{*}{Max Pool} & Natural && \textbf{85.7} & \textbf{86.1} && 0.0 & 0.0 \\
    & Adversarial && 84.4 & 85.6 && 22.2 & 3.2 \\
    & Provable && 77.6 & 79.1 && \textbf{91.4} & \textbf{84.7} \\
    \addlinespace
    \multirow{3}{*}{Average Pool} & Natural && 81.9 & 84.0 && 0.0 & 0.0 \\
    & Adversarial && 80.1 & 80.7 && 47.7 & 43.7 \\
    & Provable && 72.7 & 72.6 && 85.3 & 76.7 \\
    \bottomrule
\end{tabular}

%% file: tables/segmentation.tex
\begin{tabular}{@{}lcrrrr@{}}
    \toprule
    $\pm \theta$ && Interval & Mohapatra \etal~\cite{mohapatra2020towards} & DeepG3D & Taylor3D \\
    \midrule
    5\textdegree && 44.1 & 50.5 & \textbf{96.6} & 96.5 \\
    10\textdegree && 32.2 & 46.5 & \textbf{95.7} & 95.6 \\
    \bottomrule
\end{tabular}

%% file: sections/conclusion.tex
\section{Conclusion}

We presented 3DCertify, the first scalable verifier able to certify robustness of 3D point cloud models against a wide range of semantic 3D transformations including rotations, shearing, twisting and tapering. The key insight of 3DCertify is a novel method which enables efficient and precise computation of linear relaxations for these transformations. Combined with an improved relaxation for max pool layers, our extensive evaluation on two datasets with object classification and part segmentation illustrates the effectiveness and broad applicability of 3DCertify.

\paragraph{Acknowledgements}

This work was partially supported by the ERC Starting Grant 680358.
We thank the anonymous reviewers for their valuable feedback.

%% file: appendix/3d_transformations.tex
\section{3D Transformations and Their Relaxations}
\label{app:transformations}

In this section, we define the semantic transformations that we consider
(\cref{app:semantic-transformations}) and provide the taylor relaxations for the
transformations (\cref{app:taylor_relaxations}), as well as their Jacobians for
DeepG3D (\cref{app:jacobians}).
We also give some background on the interval arithmetic used to compute bounds on
the approximation error (\cref{app:interval_arithmetic}).

\subsection{Semantic Transformations}
\label{app:semantic-transformations}

3DCertify can handle a wide range of semantic transformations, including 3D rotation
around any axis with $\theta \in \mathbb{R}$ as defined in \cref{eq:rotation-z}.
We can also certify shearing, twisting, and tapering of a point cloud,
defined pointwise (since each point is transformed independently) for a point
$\bm{p} = (x, y, z)^T$ as
\begin{gather*}
	\mathrm{Shear}(\bm{p}, \bm{\theta}) = \begin{pmatrix}
		\theta_1 z + x \\
        \theta_2 z + y \\
        z
	\end{pmatrix} \\
	\mathrm{Twist}(\bm{p}, \theta) = \begin{pmatrix}
		x \cos(\theta z) - y \sin(\theta z) \\
        x \sin(\theta z) + y \cos(\theta z) \\
        z
	\end{pmatrix} \\
	\mathrm{Taper}(\bm{p}, \bm{\theta}) = \begin{pmatrix}
		(\frac{1}{2} \theta_1^2 z + \theta_2 z + 1) x \\
		(\frac{1}{2} \theta_1^2 z + \theta_2 z + 1) y \\
		z
	\end{pmatrix}
\end{gather*}

or any composition of these functions.

\subsection{Taylor Relaxations}
\label{app:taylor_relaxations}

In \cref{sec:taylor}, we presented the general form of our linear bounds $f_l(P, \bm{\theta})$
and $f_u(P, \bm{\theta})$ for any twice continuously differentiable transformation function
$f(P, \bm{\theta})$, as well as the relaxation for $\mathrm{Rot}_Z$ as an example. Rotation around
the other two axes, \ie, $\mathrm{Rot}_X$ and $\mathrm{Rot}_Y$ can be computed analogously.
Here, we list
the linear relaxations for the remaining transformation functions we use in our experiments.

All transformations can be applied to each point individually, allowing us to denote them
for a single point as
$f(\bm{p}, \bm{\theta})$ with $\bm{p} = (x, y, z)^T \in P$. For each transformation, we list the
first-order taylor polynomial $Q(\bm{p}, \bm{\theta})$ and remainder $R(\bm{p}, \bm{\theta})$,
such that $f(\bm{p}, \bm{\theta}) = Q(\bm{p}, \bm{\theta}) + R(\bm{p}, \bm{\theta})$. As described
in \cref{sec:taylor}, we use interval arithmetic (\cref{app:interval_arithmetic}) to get real-valued
bounds $\bm{l}_R \leq R(\bm{p}, \bm{\bar{\theta}}) \leq \bm{u}_R$ for the interval
$\bm{\bar{\theta}} = [\bm{l_\theta}, \bm{u_\theta}]$ with $\bm{t} = (\bm{l_\theta} + \bm{u_\theta})/2$
and therefore the lower constraint
$f_l(\bm{p}, \bm{\theta}) = Q(\bm{p}, \bm{\theta}) + \bm{l}_R$ and upper constraint
$f_u(\bm{p}, \bm{\theta}) = Q(\bm{p}, \bm{\theta}) + \bm{u}_R$.

\noindent Shearing:
\begin{align*}
    &Q_{\mathrm{Shear}}(\bm{p}, \bm{\theta})~
    \begin{aligned}[t]
    &= \begin{pmatrix}
    t_1 z + x \\
    t_2 z + y \\
    z
    \end{pmatrix}\\
    &+ \begin{pmatrix}
    z \\
    0 \\
    0 \\
    \end{pmatrix} (\theta_1 - t_1)  + \begin{pmatrix}
    0 \\
    z \\
    0 \\
    \end{pmatrix} (\theta_2 - t_2)
    \end{aligned} \\
    &R_{\mathrm{Shear}}(\bm{p}, \bm{\bar{\theta}}) = 0
\end{align*}
Twisting:
\begin{align*}
    &Q_{\mathrm{Twist}}(\bm{p}, \theta)~
    \begin{aligned}[t]
    = &\begin{pmatrix}
    \cos(tz)x - \sin(tz)y \\
    \sin(tz)x + \cos(tz)y \\
    z
    \end{pmatrix} \\
    + &\begin{pmatrix}
    	-z(\sin(tz)x + \cos(tz)y) \\
    	z(\cos(\theta z)x - \sin(\theta z)y) \\
    	0 \\
    \end{pmatrix} (\theta - t)
    \end{aligned} \\
    &R_{\mathrm{Twist}}(\bm{p}, \bar{\theta}) = \frac{1}{2} \begin{pmatrix}
    	-z^2(\cos(\bar{\theta} z)x - \sin(\bar{\theta} z)y) \\
    	-z^2(\sin(\bar{\theta} z)x + \cos(\bar{\theta} z)y) \\
    	0
    \end{pmatrix} (\bar{\theta} - t)^2
\end{align*}
Tapering:
\begin{align*}
    &Q_{\mathrm{Taper}}(\bm{p}, \bm{\theta})~
    \begin{aligned}[t]
    = &\begin{pmatrix}
    (\frac{1}{2} t_1^2 z + t_2 z + 1) x \\
    (\frac{1}{2} t_1^2 z + t_2 z + 1) y \\
    z
    \end{pmatrix} \\
    + &\begin{pmatrix}
    	t_1 z x \\
    	t_1 z y \\
    	0 \\
    \end{pmatrix} (\theta_1 - t_1)
    + \begin{pmatrix}
    	z x \\
    	z y \\
    	0 \\
    \end{pmatrix} (\theta_2 - t_2)
    \end{aligned} \\
    &R_{\mathrm{Taper}}(\bm{p}, \bm{\bar{\theta}}) = \frac{1}{2} \begin{pmatrix}
    	z x \\
    	z y \\
    	0
    \end{pmatrix} (\bar{\theta_1} - t_1)^2
\end{align*}

\subsection{Interval Arithmetic}
\label{app:interval_arithmetic}

When evaluating functions such as $R(P, \bm{\theta})$ on intervals, we use the following standard operators:
\begin{align*}
	&- [x_l, x_u] = [-x_u, -x_l] \\
	&[x_l, x_u] + [y_l, y_u] = [x_l + y_l, x_u + y_u] \\
	&[x_l, x_u] - [y_l, y_u] = [x_l - y_u, x_u - y_l] \\
	&[x_l, x_u] \cdot [y_l, y_u] = \begin{aligned}[t][&\min(x_l y_l, x_l y_u, x_u y_l, x_u y_u), \\
	&\max(x_l y_l, x_l y_u, x_u y_l, x_u y_u)]\end{aligned}
\end{align*}
For mixed operations with scalars, \ie, $a * [x_l, x_u]$, we can treat the scalar as an interval
with one element: $[a, a] * [x_l, x_u]$.

In addition to these basic operators, we use the following sine function:
\begin{align*}
	&\sin([x_l, x_u]) = [y_l, y_u]\text{, where} \\
	&\quad y_l = \begin{cases}
		-1 & -\frac{\pi}{2} + 2k\pi \in [x_l, x_u] \\
		\min(\sin(x_l), \sin(x_u)) & \text{otherwise}
	\end{cases} \\
		&\quad y_u = \begin{cases}
		1 & \frac{\pi}{2} + 2k\pi \in [x_l, x_u] \\
		\max(\sin(x_l), \sin(x_u)) & \text{otherwise}
	\end{cases}
\end{align*}
with $k \in \mathbb{Z}.$ Similarly, we can define the cosine function:
\begin{align*}
	&\cos([x_l, x_u]) = [y_l, y_u]\text{, where} \\
	&\quad y_l = \begin{cases}
		-1 & \pi + 2k\pi \in [x_l, x_u] \\
		\min(\cos(x_l), \cos(x_u)) & \text{otherwise}
	\end{cases} \\
		&\quad y_u = \begin{cases}
		1 & 2k\pi \in [x_l, x_u] \\
		\max(\cos(x_l), \cos(x_u)) & \text{otherwise}.
	\end{cases}
\end{align*}

To compute the square $x^2$ of $x = [x_l, x_u]$, we could simply use $x \cdot x$. While sound,
we can compute a tighter interval for some cases:
\begin{equation*}
	[x_l, x_u]^2 = \begin{cases}
		[x_l^2, x_u^2] &x_l \geq 0 \\
		[x_u^2, x_l^2] &x_u \leq 0 \\
		[0, \max(x_l^2, x_u^2)] & \text{otherwise}.
	\end{cases}
\end{equation*}

Using these operators and functions, we can evaluate all of our relaxations and
their derivatives with intervals as input.

\subsection{Jacobian Matrices}
\label{app:jacobians}

Computing linear relaxations using Deepg3D (\cref{sec:deepg}) requires the Jacobians of our
3D transformations, both with respect to the transformation parameters and with respect to
the point cloud inputs.
For example, for 3D rotation around the $z$-axis with $\theta \in \mathbb{R}$, as
defined in \cref{eq:rotation-z}, we compute
\begin{equation}
    \partial_{\bm{p}} \mathrm{Rot}_Z(\bm{p}, \theta) = \begin{pmatrix}
    \cos(\theta) & -\sin(\theta) & 0 \\
    \sin(\theta) & \cos(\theta)  & 0 \\
    0            & 0             & 1
    \end{pmatrix},
    \label{eq:rotationz_dp}
\end{equation}
and
\begin{equation}
    \partial_\theta \mathrm{Rot}_Z(\bm{p}, \theta) = \begin{pmatrix}
    -x \sin(\theta) - y \cos(\theta) \\
    x \cos(\theta)  - y \sin(\theta) \\
    z
    \end{pmatrix}.
    \label{eq:rotationz_dt}
\end{equation}
The corresponding Jacobians for $\mathrm{Shear}$, $\mathrm{Twist}$ and
$\mathrm{Taper}$ (\cref{sec:experiments_transformations}) are given by:
\begin{align*}
    \partial_{\bm{p}} \mathrm{Shear}(\bm{p}, \bm{\theta}) =& \begin{pmatrix}
    1 & 0 & \theta_1 \\
    0 & 1 & \theta_2 \\
    0 & 0 & 1
    \end{pmatrix} \\
    \partial_{\bm{\theta}} \mathrm{Shear}(\bm{p}, \bm{\theta}) =& \begin{pmatrix}
    z & 0 \\
    0 & z \\
    0 & 0
    \end{pmatrix}
\end{align*}
\resizebox{.95\columnwidth}{!}{
\begin{minipage}{\linewidth}
\begin{align*}
	\partial_{\bm{p}} \mathrm{Twist}(\bm{p}, \theta) =& \begin{pmatrix}
	\cos(\theta z) & -sin(\theta z) & -\theta (\sin(\theta z) x + \cos(\theta z) y) \\
	\sin(\theta z) & \cos(\theta z) & \theta(\cos(theta z) x - \sin(\theta z) y) \\
	0			   & 0				& 1
	\end{pmatrix} \\
	\partial_{\theta} \mathrm{Twist}(\bm{p}, \theta) =& \begin{pmatrix}
	-z (\sin(\theta z) x + \cos(\theta z) y) \\
	z (\cos(\theta z) x - \sin(\theta z) y) \\
	0
	\end{pmatrix}
\end{align*}
\end{minipage}
}
\resizebox{.95\columnwidth}{!}{
\begin{minipage}{\linewidth}
\begin{align*}
	\partial_{\bm{p}} \mathrm{Taper}(\bm{p}, \bm{\theta}) =& \begin{pmatrix}
	\frac{1}{2} \theta_1^2 z + \theta_2 z + 1 & 0 & (\frac{1}{2} \theta_1^2 + \theta_2) x  \\
	0 & \frac{1}{2} \theta_1^2 z + \theta_2 z & (\frac{1}{2} \theta_1^2 + \theta_2) y \\
	0 & 0 & 1
	\end{pmatrix} \\
	\partial_{\bm{\theta}} \mathrm{Taper}(\bm{p}, \bm{\theta}) =& \begin{pmatrix}
	\theta_1 z x & z x \\
	\theta_1 z y & z y \\
	0 & 0
	\end{pmatrix}
\end{align*}
\end{minipage}
}

%% file: appendix/proof_composition.tex
\section{Proof for Composition of Transformations}
\label{app:composition}

To show that our taylor approximations introduced in \cref{sec:taylor} can be
applied to the composition of multiple transformations, we show that the
composition of two twice continuously differentiable functions is itself twice
continuously differentiable.
That is, given two twice continuously differentiable functions
$f: \mathbb{R}^n \mapsto \mathbb{R}^p$ and
$g: \mathbb{R}^m \mapsto \mathbb{R}^n$, we want to show that $h = f \circ g$ is
also twice continuously differentiable.

To simplify notation, we define $\bm{y} = f(\bm{u})$ and $\bm{u} = g(\bm{x})$.
Using the chain rule, we know that the first-order derivatives exist and can,
with slight liberties in notation, be written as:
\begin{equation}
	\frac{\partial \bm{y}}{\partial x_i} = \sum_k{\frac{\partial \bm{y}}{\partial u_k} \frac{\partial u_k}{\partial x_i}}.
	\label{eq:first_order_derivatives}
\end{equation}
Furthermore, we know that $f$ and $g$ are twice differentiable, hence
\cref{eq:first_order_derivatives} consists of compositions, products and sums of
differentiable functions and thus is again differentiable.
We therefore conclude that $f \circ g$ is itself twice differentiable.

It remains to be shown that the second-order derivatives are continuous.
Using Fa\`{a} di Bruno's formula, we can write the second-order derivatives as:
\begin{equation}
	\frac{\partial^2\bm{y}}{\partial x_i \partial x_j} =
		\sum_k{\frac{\partial y}{\partial u_k}\frac{\partial^2 u_k}{\partial x_i \partial x_j}} +
		\sum_k{\sum_l{ \frac{\partial^2 \bm{y}}{\partial u_k \partial u_l}
		\frac{\partial u_k}{\partial x_i} \frac{\partial u_l}{\partial x_j}}}.
	\label{eq:second_order_derivatives}
\end{equation}
We know that $f$, $g$ and their first- and second-order derivatives are
continuous.
\Cref{eq:second_order_derivatives} is therefore a combination of compositions,
products and sums of continuous functions, which means it is itself continuous.
This means $f \circ g$ is twice continuously differentiable and we can therefore
calculate Taylor bounds for any composition of twice continuously differentiable
transformations.

%% file: appendix/pointnet_architectures.tex
\section{PointNet Architectures}
\label{app:architectures}

For both object classification and part segmentation, we use
PointNet~\cite{qi2017pointnet} models.
Below we present the exact layer configurations used.

\paragraph{Object Classification} \mbox{} \\

For object classification, we use the following network architecture:
\\\\
\noindent \begin{tabular*}{\columnwidth}{@{}l@{\extracolsep{\fill}}llll@{}}
	No\phantom{ab} & Type & Normalization & Activation & Features \\
	\midrule
	1 & Linear & BatchNorm & ReLU & 64 \\
	2 & Linear & BatchNorm & ReLU & 64 \\
	3 & Linear & BatchNorm & ReLU & 64 \\
	4 & Linear & BatchNorm & ReLU & 128 \\
	5 & Linear & BatchNorm & ReLU & 1024 \\
	\addlinespace
	6 & MaxPool & & & 1024 \\
	\addlinespace
	7 & Linear & BatchNorm & ReLU & 512 \\
	8 & Linear & BatchNorm & ReLU & 256 \\
	9 & Linear & & SoftMax & num classes \\
\end{tabular*}
\\\\
The first block of linear (fully connected) layers (no 1 to 5) is executed on
each point individually, but sharing weights across all points.
We implement this via 1D convolution layers with stride 1 as in the original
work by Qi \etal~\cite{qi2017pointnet}.
Layer 6 pools each feature across points.
During training, a dropout of 30\% is applied for layer 8.

\paragraph{Part Segmentation} \mbox{} \\

\noindent \begin{tabular*}{\columnwidth}{@{}@{\extracolsep{\fill}}lllll@{}}
	No & Type & Normalization & Activation & Features \\
	\midrule
	1 & Linear & BatchNorm & ReLU & 64 \\
	2 & Linear & BatchNorm & ReLU & 128 \\
	3 & Linear & BatchNorm & ReLU & 256 \\
	4 & Linear & BatchNorm & & 128 \\
	\addlinespace
	5 & MaxPool & & & 128 \\
	6 & Repeat & & & 128 \\
	7 & \multicolumn{3}{l}{Concatenate (1, 2, 3, 6)} & 576 \\
	\addlinespace
	8 & Linear & BatchNorm & ReLU & 256 \\
	9 & Linear & BatchNorm & ReLU & 128 \\
	10 & Linear & & SoftMax & num parts \\
\end{tabular*}
\\\\

The architecture for part segmentation differs in some ways, since it needs to
predict a label for each point individually.
Again, the first block of linear layers (1-4) is applied to each point
individually with shared weights and max pool combines per-point features to one
global feature vector.
Layer 7 concatenates the local features of layers 1 to 3 with the global feature
from layer 5 for each point by simple concatenation.
The last 3 linear layers are again applied individually for each point on the
combined local and global feature and predict the part the particular point
belongs to.

Since DeepPoly~\cite{singh2019abstract} cannot handle this architecture for part
segmentation, we implement novel relaxations for the concatenation and repeat
layers.
In particular, the transformer for concatenation requires the verifier to handle
layers with multiple predecessors, which is out of reach for current
state-of-the-art verifiers.
We provide our implementation in the accompanying code.

%% file: appendix/additional_experiments.tex
\section{Additional Experiments}
\label{app:additional-experiments}

In this section, we present additional empirical evidence for the benefits of
our improved max pool relaxations in \cref{app:max-pool}, and investigate the
effect different max pool group sizes have on certification results.
We also show that Taylor3D efficiently scales to real-world point cloud sizes in
\cref{app:scaling}, with running times of only a few milliseconds.

\subsection{Improved Max Pool Relaxation}
\label{app:max-pool}

\begin{table}
    \begin{center}
    \resizebox{\columnwidth}{!}{ \input{tables/maxpool_cv} }
    \end{center}
    \caption{
        Percentage of certified images with different max pool relaxations
        for two different image classification tasks for $\ell_{\infty}$
        noise perturbations.
    }
    \label{tbl:maxpool_cv}
\end{table}

\paragraph*{Applications beyond point clouds}

In \cref{sec:experiments_maxpool}, we show that our improved max pool
relaxation, introduced in \cref{sec:maxpool}, significantly improves
certification for PointNet models compared to the previous state-of-the-art,
especially for models with larger pooling layers.
Here, we demonstrate that our new relaxations are useful beyond PointNet, \ie, for
any network architecture containing max pool layers.
To that end, we show certification results for a convolutional image
classification model with nine conv/linear layers and two max pool layers for
the MNIST~\cite{lecun1998gradient} dataset in \cref{tbl:maxpool_cv}, comparing
our improved relaxations with the best baseline.
Our improved max pool relaxations consistently outperform the previous
state-of-the-art across all $\epsilon$-values, significantly
increasing the number of images for which we can certify correct classification.
These results demonstrate that models beyond the 3D point cloud domain benefit
from our new relaxations.

\begin{table}
    \begin{center}
    \input{tables/maxpool_group_size}
    \end{center}
    \caption{
        Percentage of certified point clouds with 64 points for different max pool
        group sizes for $\pm3\degree$ rotation.
    }
    \label{tbl:maxpool_group_size}
\end{table}

\paragraph{Max pool group size}

Our improved max pool relaxation, introduced in \cref{sec:maxpool}, requires computing
the convex hull of the polyhedral relaxation, for which the running time grows exponentially
in the number of input neurons. This is why we recursively split the max pool operation
into sub groups. \cref{tbl:maxpool_group_size} shows the certification accuracy and
average running time of DeepPoly with the improved max pool relaxation for different group
sizes.
Increasing the group size beyond this range is impractical (\ie, more than 3h
per point clouds) due to the exponential scaling behavior of convex hull
computation.
Nevertheless, our experiments indicate that our recursively partitioned
relaxation is not impeded by this constraint since the different group sizes do
not influence certification performance (while, in theory, computing the
relaxation over all inputs should be most precise), allowing us to optimize for
improved running time.

\subsection{Scaling}
\label{app:scaling}

\begin{table}
    \begin{center}
    \resizebox{\columnwidth}{!}{ \input{tables/scaling} }
    \end{center}
    \caption{
        Running time in seconds to compute the relaxations for different
        real-world point cloud sizes with Taylor3D and DeepG3D.
        Taylor3D achieves speed-ups of up to 14\,442x for $\mathrm{Rot}_{Z}$ and
        5\,624x for $\mathrm{Twist}$.
    }
    \label{tbl:scaling}
\end{table}

3D processing of LIDAR point cloud data is an active area of research.
The main challenge is the huge size of point clouds (in the order of 100k
points~\cite{geiger2013vision}) that have to be processed in real-time for most
applications.
Both DeepG3D and Taylor3D scale linearly with point cloud size and can be parallelized
perfectly across points. \cref{tbl:scaling} shows the running time of computing linear
relaxations for large point cloud sizes using Taylor3D and Deepg3D respectively.
All experiments are run on an AMD EPYC 7601 processor with 2.2 GHz. Taylor3D is
efficiently implemented as vectorized operations using Numpy~\cite{harris2020array}
and run on a single thread. DeepG3D uses the original parallelized
implementation~\cite{balunovic2019certifying} and runs in parallel with 16
threads.
The results show that, while both implementations scale linearly in the point
cloud size, Taylor3D is significantly more efficient, computing relaxations in
only a few milliseconds event for large point cloud sizes on a single thread,
thereby achieving speed-ups of up to 14\,442x over DeepG3D.
This enables easy and efficient scaling to real-world applications.

%% file: tables/maxpool_cv.tex
\begin{tabularx}{\columnwidth}{@{}Xcrcrcr@{}}
    \toprule
    $\epsilon$ && \multicolumn{1}{c}{0.005} && \multicolumn{1}{c}{0.010} && \multicolumn{1}{c@{}}{0.015} \\
    \midrule
    DeepPoly~\cite{singh2019abstract} && 72.8 && 33.3 && 3.7 \\
    Ours && \textbf{79.0} && \textbf{38.3} && \textbf{6.2} \\
    \bottomrule
\end{tabularx}

%% file: tables/maxpool_group_size.tex
\begin{tabularx}{\columnwidth}{@{}Xcrrr@{}}
    \toprule
    Group Size && Certified (\%) & Time (s) \\
    \midrule
    4 && 93.5 & 56 \\
    8 && 93.5 & 76 \\
    12 && 93.5 & 119 \\
    \bottomrule
\end{tabularx}

%% file: tables/scaling.tex
\begin{tabular}{@{}lcrrcrr@{}}
    \toprule
    && \multicolumn{2}{c}{$\mathrm{Rot}_{Z}$} && \multicolumn{2}{c}{$\mathrm{Twist}$} \\
    \cmidrule{3-4}
    \cmidrule{6-7}
    Points && Taylor3D & DeepG3D && Taylor3D & DeepG3D \\
    \midrule
    100\,000 && 0.036 & 393 && 0.070 & 366 \\
    200\,000 && 0.070 & 887 && 0.169 & 848 \\
    300\,000 && 0.104 & 1502 && 0.266 & 1496 \\
    \bottomrule
\end{tabular}

%% file: appendix/maxpool_analysis.tex
\section{Max Pool Analysis}
\label{app:maxpool_analysis}

\begin{figure}
	\includegraphics[width=\columnwidth]{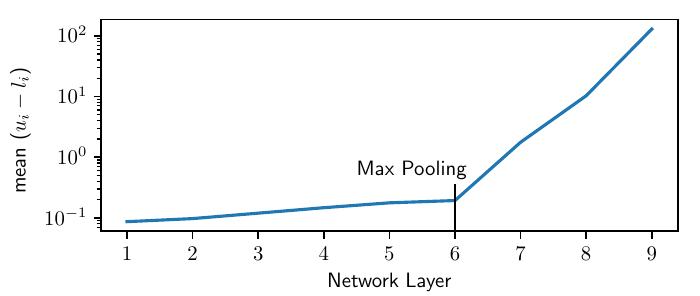}
	\vspace{-4mm}
	\caption{
        Plotting the mean difference between upper and lower bounds of neurons
        after each layer shows that the precision significantly decreases after
        the max pool and therefore motivates the need for improvement.
        Note the logarithmic scaling of the y-axis.
    }
	\label{fig:maxpool}
	\vspace{-5mm}
\end{figure}

The state-of-the-art linear relaxations for max pool are imprecise,
especially for many inputs to the pooling layer.
We demonstrate this by plotting the mean divergence between DeepPoly's upper and
lower bounds for each of PointNet's layers in \cref{fig:maxpool}, where we
observe that the bounds start to significantly diverge after the max pool
layer.

%% file: main.bbl
\begin{thebibliography}{10}\itemsep=-1pt

\bibitem{athalye2018obfuscated}
Anish Athalye, Nicholas Carlini, and David~A. Wagner.
\newblock Obfuscated gradients give a false sense of security: Circumventing
  defenses to adversarial examples.
\newblock In {\em Proceedings of the 35th International Conference on Machine
  Learning}, 2018.

\bibitem{balunovic2019certifying}
Mislav Balunovic, Maximilian Baader, Gagandeep Singh, Timon Gehr, and Martin~T.
  Vechev.
\newblock Certifying geometric robustness of neural networks.
\newblock In {\em Advances in Neural Information Processing Systems 32}, 2019.

\bibitem{balunovic2020adversarial}
Mislav Balunovic and Martin~T. Vechev.
\newblock Adversarial training and provable defenses: Bridging the gap.
\newblock In {\em 8th International Conference on Learning Representations},
  2020.

\bibitem{bonaert2021fast}
Gregory Bonaert, Dimitar~I. Dimitrov, Maximilian Baader, and Martin~T. Vechev.
\newblock Fast and precise certification of transformers.
\newblock In {\em 42nd {ACM} {SIGPLAN} International Conference on Programming
  Language Design and Implementation}, 2021.

\bibitem{boopathy2019cnncert}
Akhilan Boopathy, Tsui{-}Wei Weng, Pin{-}Yu Chen, Sijia Liu, and Luca Daniel.
\newblock Cnn-cert: An efficient framework for certifying robustness of
  convolutional neural networks.
\newblock In {\em The Thirty-Third {AAAI} Conference on Artificial
  Intelligence}, 2019.

\bibitem{cao2019adversarial}
Yulong Cao, Chaowei Xiao, Dawei Yang, Jing Fang, Ruigang Yang, Mingyan Liu, and
  Bo Li.
\newblock Adversarial objects against lidar-based autonomous driving systems.
\newblock {\em CoRR}, abs/1907.05418, 2019.

\bibitem{chang2015shapenet}
Angel~X. Chang, Thomas~A. Funkhouser, Leonidas~J. Guibas, Pat Hanrahan,
  Qi{-}Xing Huang, Zimo Li, Silvio Savarese, Manolis Savva, Shuran Song, Hao
  Su, Jianxiong Xiao, Li Yi, and Fisher Yu.
\newblock Shapenet: An information-rich 3d model repository.
\newblock {\em CoRR}, abs/1512.03012, 2015.

\bibitem{chen20213d}
Siheng Chen, Baoan Liu, Chen Feng, Carlos Vallespi{-}Gonzalez, and Carl
  Wellington.
\newblock 3d point cloud processing and learning for autonomous driving:
  Impacting map creation, localization, and perception.
\newblock {\em {IEEE} Signal Process. Mag.}, 38, 2021.

\bibitem{chen2017multiview}
Xiaozhi Chen, Huimin Ma, Ji Wan, Bo Li, and Tian Xia.
\newblock Multi-view 3d object detection network for autonomous driving.
\newblock In {\em {IEEE} Conference on Computer Vision and Pattern
  Recognition}, 2017.

\bibitem{rosenfeld2019certified}
Jeremy~M. Cohen, Elan Rosenfeld, and J.~Zico Kolter.
\newblock Certified adversarial robustness via randomized smoothing.
\newblock In {\em Proceedings of the 36th International Conference on Machine
  Learning}, 2019.

\bibitem{fischer2020certification}
Marc Fischer, Maximilian Baader, and Martin~T. Vechev.
\newblock Certified defense to image transformations via randomized smoothing.
\newblock In {\em Advances in Neural Information Processing Systems 33}, 2020.

\bibitem{fischer2021scalable}
Marc Fischer, Maximilian Baader, and Martin~T. Vechev.
\newblock Scalable certified segmentation via randomized smoothing.
\newblock In {\em Proceedings of the 38th International Conference on Machine
  Learning}, 2021.

\bibitem{fukuda1995double}
Komei Fukuda and Alain Prodon.
\newblock Double description method revisited.
\newblock In {\em Combinatorics and Computer Science}, 1995.

\bibitem{gehr2018ai2}
Timon Gehr, Matthew Mirman, Dana Drachsler{-}Cohen, Petar Tsankov, Swarat
  Chaudhuri, and Martin~T. Vechev.
\newblock {AI2:} safety and robustness certification of neural networks with
  abstract interpretation.
\newblock In {\em {IEEE} Symposium on Security and Privacy}, 2018.

\bibitem{geiger2013vision}
Andreas Geiger, Philip Lenz, Christoph Stiller, and Raquel Urtasun.
\newblock Vision meets robotics: The {KITTI} dataset.
\newblock {\em Int. J. Robotics Res.}, 32, 2013.

\bibitem{goodfellow2015explaining}
Ian~J. Goodfellow, Jonathon Shlens, and Christian Szegedy.
\newblock Explaining and harnessing adversarial examples.
\newblock In {\em 3rd International Conference on Learning Representations},
  2015.

\bibitem{gowal2019scalable}
Sven Gowal, Krishnamurthy Dvijotham, Robert Stanforth, Rudy Bunel, Chongli Qin,
  Jonathan Uesato, Relja Arandjelovic, Timothy~Arthur Mann, and Pushmeet Kohli.
\newblock Scalable verified training for provably robust image classification.
\newblock In {\em {IEEE/CVF} International Conference on Computer Vision}.
  {IEEE}, 2019.

\bibitem{harris2020array}
Charles~R. Harris, K.~Jarrod Millman, St{'{e}}fan~J. van~der Walt, Ralf
  Gommers, Pauli Virtanen, David Cournapeau, Eric Wieser, Julian Taylor,
  Sebastian Berg, Nathaniel~J. Smith, Robert Kern, Matti Picus, Stephan Hoyer,
  Marten~H. van Kerkwijk, Matthew Brett, Allan Haldane, Jaime~Fern{'{a}}ndez
  del R{'{\i}}o, Mark Wiebe, Pearu Peterson, Pierre G{'{e}}rard-Marchant, Kevin
  Sheppard, Tyler Reddy, Warren Weckesser, Hameer Abbasi, Christoph Gohlke, and
  Travis~E. Oliphant.
\newblock Array programming with {NumPy}.
\newblock {\em Nature}, 585(7825):357--362, Sept. 2020.

\bibitem{katz2017reluplex}
Guy Katz, Clark~W. Barrett, David~L. Dill, Kyle Julian, and Mykel~J.
  Kochenderfer.
\newblock Reluplex: An efficient {SMT} solver for verifying deep neural
  networks.
\newblock In {\em Computer Aided Verification - 29th International Conference},
  2017.

\bibitem{lang2020geometric}
Itai Lang, Uriel Kotlicki, and Shai Avidan.
\newblock Geometric adversarial attacks and defenses on 3d point clouds.
\newblock {\em CoRR}, abs/2012.05657, 2020.

\bibitem{lecun1998gradient}
Yann LeCun, L{\'e}on Bottou, Yoshua Bengio, and Patrick Haffner.
\newblock Gradient-based learning applied to document recognition.
\newblock {\em Proceedings of the IEEE}, 86(11):2278--2324, 1998.

\bibitem{lecuyer2019certified}
Mathias L{\'{e}}cuyer, Vaggelis Atlidakis, Roxana Geambasu, Daniel Hsu, and
  Suman Jana.
\newblock Certified robustness to adversarial examples with differential
  privacy.
\newblock In {\em {IEEE} Symposium on Security and Privacy}, 2019.

\bibitem{lee2020shapeadv}
Kibok Lee, Zhuoyuan Chen, Xinchen Yan, Raquel Urtasun, and Ersin Yumer.
\newblock Shapeadv: Generating shape-aware adversarial 3d point clouds.
\newblock {\em CoRR}, abs/2005.11626, 2020.

\bibitem{li2020provable}
Linyi Li, Maurice Weber, Xiaojun Xu, Luka Rimanic, Tao Xie, Ce Zhang, and Bo
  Li.
\newblock Provable robust learning based on transformation-specific smoothing.
\newblock {\em CoRR}, abs/2002.12398, 2020.

\bibitem{li2020rotation}
Xianzhi Li, Ruihui Li, Guangyong Chen, Chi{-}Wing Fu, Daniel Cohen{-}Or, and
  Pheng{-}Ann Heng.
\newblock A rotation-invariant framework for deep point cloud analysis.
\newblock {\em CoRR}, abs/2003.07238, 2020.

\bibitem{liang2018deep}
Ming Liang, Bin Yang, Shenlong Wang, and Raquel Urtasun.
\newblock Deep continuous fusion for multi-sensor 3d object detection.
\newblock In {\em Computer Vision - 15th European Conference}, 2018.

\bibitem{lin2019robustness}
Wang Lin, Zhengfeng Yang, Xin Chen, Qingye Zhao, Xiangkun Li, Zhiming Liu, and
  Jifeng He.
\newblock Robustness verification of classification deep neural networks via
  linear programming.
\newblock In {\em {IEEE} Conference on Computer Vision and Pattern
  Recognition}, 2019.

\bibitem{liu2019extending}
Daniel Liu, Ronald Yu, and Hao Su.
\newblock Extending adversarial attacks and defenses to deep 3d point cloud
  classifiers.
\newblock In {\em {IEEE} International Conference on Image Processing}, 2019.

\bibitem{liu2020adversarial}
Daniel Liu, Ronald Yu, and Hao Su.
\newblock Adversarial shape perturbations on 3d point clouds.
\newblock In {\em Computer Vision - {ECCV} Workshops}, 2020.

\bibitem{liu2021pointguard}
Hongbin Liu, Jinyuan Jia, and Neil~Zhenqiang Gong.
\newblock Pointguard: Provably robust 3d point cloud classification.
\newblock {\em CoRR}, abs/2103.03046, 2021.

\bibitem{madry2018towards}
Aleksander Madry, Aleksandar Makelov, Ludwig Schmidt, Dimitris Tsipras, and
  Adrian Vladu.
\newblock Towards deep learning models resistant to adversarial attacks.
\newblock In {\em 6th International Conference on Learning Representations},
  2018.

\bibitem{mirman2018differentiable}
Matthew Mirman, Timon Gehr, and Martin~T. Vechev.
\newblock Differentiable abstract interpretation for provably robust neural
  networks.
\newblock In {\em Proceedings of the 35th International Conference on Machine
  Learning}, 2018.

\bibitem{mohapatra2020towards}
Jeet Mohapatra, Tsui{-}Wei Weng, Pin{-}Yu Chen, Sijia Liu, and Luca Daniel.
\newblock Towards verifying robustness of neural networks against {A} family of
  semantic perturbations.
\newblock In {\em {IEEE/CVF} Conference on Computer Vision and Pattern
  Recognition}, 2020.

\bibitem{muller2021precise}
Mark~Niklas M{\"{u}}ller, Gleb Makarchuk, Gagandeep Singh, Markus
  P{\"{u}}schel, and Martin~T. Vechev.
\newblock Precise multi-neuron abstractions for neural network certification.
\newblock {\em CoRR}, abs/2103.03638, 2021.

\bibitem{pei2017towards}
Kexin Pei, Yinzhi Cao, Junfeng Yang, and Suman Jana.
\newblock Towards practical verification of machine learning: The case of
  computer vision systems.
\newblock {\em CoRR}, abs/1712.01785, 2017.

\bibitem{qi2017pointnet}
Charles~Ruizhongtai Qi, Hao Su, Kaichun Mo, and Leonidas~J. Guibas.
\newblock Pointnet: Deep learning on point sets for 3d classification and
  segmentation.
\newblock In {\em {IEEE} Conference on Computer Vision and Pattern
  Recognition}, 2017.

\bibitem{qi2017pointnet++}
Charles~Ruizhongtai Qi, Li Yi, Hao Su, and Leonidas~J. Guibas.
\newblock Pointnet++: Deep hierarchical feature learning on point sets in a
  metric space.
\newblock In {\em Advances in Neural Information Processing Systems 30}, 2017.

\bibitem{raghunathan2018semidefinite}
Aditi Raghunathan, Jacob Steinhardt, and Percy Liang.
\newblock Semidefinite relaxations for certifying robustness to adversarial
  examples.
\newblock In {\em Advances in Neural Information Processing Systems 31}, 2018.

\bibitem{ruoss2020efficient}
Anian Ruoss, Maximilian Baader, Mislav Balunovic, and Martin~T. Vechev.
\newblock Efficient certification of spatial robustness.
\newblock In {\em Thirty-Fifth {AAAI} Conference on Artificial Intelligence},
  2021.

\bibitem{salman2019provably}
Hadi Salman, Jerry Li, Ilya~P. Razenshteyn, Pengchuan Zhang, Huan Zhang,
  S{\'{e}}bastien Bubeck, and Greg Yang.
\newblock Provably robust deep learning via adversarially trained smoothed
  classifiers.
\newblock In {\em Advances in Neural Information Processing Systems 32}, 2019.

\bibitem{salman2019convex}
Hadi Salman, Greg Yang, Huan Zhang, Cho{-}Jui Hsieh, and Pengchuan Zhang.
\newblock A convex relaxation barrier to tight robustness verification of
  neural networks.
\newblock In {\em Advances in Neural Information Processing Systems 32}, 2019.

\bibitem{singh2019beyond}
Gagandeep Singh, Rupanshu Ganvir, Markus P{\"{u}}schel, and Martin~T. Vechev.
\newblock Beyond the single neuron convex barrier for neural network
  certification.
\newblock In {\em Advances in Neural Information Processing Systems 32}, 2019.

\bibitem{singh2018fast}
Gagandeep Singh, Timon Gehr, Matthew Mirman, Markus P{\"{u}}schel, and
  Martin~T. Vechev.
\newblock Fast and effective robustness certification.
\newblock In {\em Advances in Neural Information Processing Systems 31}, 2018.

\bibitem{singh2019abstract}
Gagandeep Singh, Timon Gehr, Markus P{\"{u}}schel, and Martin~T. Vechev.
\newblock An abstract domain for certifying neural networks.
\newblock {\em Proc. {ACM} Program. Lang.}, 3, 2019.

\bibitem{singh2019boosting}
Gagandeep Singh, Timon Gehr, Markus P{\"{u}}schel, and Martin~T. Vechev.
\newblock Boosting robustness certification of neural networks.
\newblock In {\em 7th International Conference on Learning Representations},
  2019.

\bibitem{sun2020adversarial}
Jiachen Sun, Karl Koenig, Yulong Cao, Qi~Alfred Chen, and Z.~Morley Mao.
\newblock On the adversarial robustness of 3d point cloud classification.
\newblock {\em CoRR}, abs/2011.11922, 2020.

\bibitem{taylor1715methodus}
Brook Taylor.
\newblock {\em Methodus incrementorum directa \& inversa / auctore Brook
  Taylor.}
\newblock Typis Pearsonianis: prostant apud Gul. Innys, 1715.

\bibitem{tjeng2019evaluating}
Vincent Tjeng, Kai~Y. Xiao, and Russ Tedrake.
\newblock Evaluating robustness of neural networks with mixed integer
  programming.
\newblock In {\em 7th International Conference on Learning Representations},
  2019.

\bibitem{tramer2020adaptive}
Florian Tram{\`{e}}r, Nicholas Carlini, Wieland Brendel, and Aleksander Madry.
\newblock On adaptive attacks to adversarial example defenses.
\newblock In {\em Advances in Neural Information Processing Systems 33}, 2020.

\bibitem{wang2018efficient}
Shiqi Wang, Kexin Pei, Justin Whitehouse, Junfeng Yang, and Suman Jana.
\newblock Efficient formal safety analysis of neural networks.
\newblock In {\em Advances in Neural Information Processing Systems 31}, 2018.

\bibitem{wang2018formal}
Shiqi Wang, Kexin Pei, Justin Whitehouse, Junfeng Yang, and Suman Jana.
\newblock Formal security analysis of neural networks using symbolic intervals.
\newblock In {\em 27th {USENIX} Security Symposium}, 2018.

\bibitem{wang2019dynamic}
Yue Wang, Yongbin Sun, Ziwei Liu, Sanjay~E. Sarma, Michael~M. Bronstein, and
  Justin~M. Solomon.
\newblock Dynamic graph {CNN} for learning on point clouds.
\newblock {\em {ACM} Trans. Graph.}, 38, 2019.

\bibitem{wen2019geometry}
Yuxin Wen, Jiehong Lin, Ke Chen, and Kui Jia.
\newblock Geometry-aware generation of adversarial and cooperative point
  clouds.
\newblock {\em CoRR}, abs/1912.11171, 2019.

\bibitem{weng2018towards}
Tsui{-}Wei Weng, Huan Zhang, Hongge Chen, Zhao Song, Cho{-}Jui Hsieh, Luca
  Daniel, Duane~S. Boning, and Inderjit~S. Dhillon.
\newblock Towards fast computation of certified robustness for relu networks.
\newblock In {\em Proceedings of the 35th International Conference on Machine
  Learning}, 2018.

\bibitem{wicker2019robustness}
Matthew Wicker and Marta Kwiatkowska.
\newblock Robustness of 3d deep learning in an adversarial setting.
\newblock In {\em {IEEE} Conference on Computer Vision and Pattern
  Recognition}, 2019.

\bibitem{wong2018provable}
Eric Wong and J.~Zico Kolter.
\newblock Provable defenses against adversarial examples via the convex outer
  adversarial polytope.
\newblock In {\em Proceedings of the 35th International Conference on Machine
  Learning}, 2018.

\bibitem{wong2020fast}
Eric Wong, Leslie Rice, and J.~Zico Kolter.
\newblock Fast is better than free: Revisiting adversarial training.
\newblock In {\em 8th International Conference on Learning Representations},
  2020.

\bibitem{wong2018scaling}
Eric Wong, Frank~R. Schmidt, Jan~Hendrik Metzen, and J.~Zico Kolter.
\newblock Scaling provable adversarial defenses.
\newblock In {\em Advances in Neural Information Processing Systems 31}, 2018.

\bibitem{wu20153d}
Zhirong Wu, Shuran Song, Aditya Khosla, Fisher Yu, Linguang Zhang, Xiaoou Tang,
  and Jianxiong Xiao.
\newblock 3d shapenets: {A} deep representation for volumetric shapes.
\newblock In {\em {IEEE} Conference on Computer Vision and Pattern
  Recognition}, 2015.

\bibitem{xiang2019generating}
Chong Xiang, Charles~R. Qi, and Bo Li.
\newblock Generating 3d adversarial point clouds.
\newblock In {\em {IEEE} Conference on Computer Vision and Pattern
  Recognition}, 2019.

\bibitem{xu2020automatic}
Kaidi Xu, Zhouxing Shi, Huan Zhang, Yihan Wang, Kai{-}Wei Chang, Minlie Huang,
  Bhavya Kailkhura, Xue Lin, and Cho{-}Jui Hsieh.
\newblock Automatic perturbation analysis for scalable certified robustness and
  beyond.
\newblock In {\em Advances in Neural Information Processing Systems 33}, 2020.

\bibitem{yang2019adversarial}
Jiancheng Yang, Qiang Zhang, Rongyao Fang, Bingbing Ni, Jinxian Liu, and Qi
  Tian.
\newblock Adversarial attack and defense on point sets.
\newblock {\em CoRR}, abs/1902.10899, 2019.

\bibitem{zhang2020towards}
Huan Zhang, Hongge Chen, Chaowei Xiao, Sven Gowal, Robert Stanforth, Bo Li,
  Duane~S. Boning, and Cho{-}Jui Hsieh.
\newblock Towards stable and efficient training of verifiably robust neural
  networks.
\newblock In {\em 8th International Conference on Learning Representations},
  2020.

\bibitem{zhang2018efficient}
Huan Zhang, Tsui{-}Wei Weng, Pin{-}Yu Chen, Cho{-}Jui Hsieh, and Luca Daniel.
\newblock Efficient neural network robustness certification with general
  activation functions.
\newblock In {\em Advances in Neural Information Processing Systems 31}, 2018.

\bibitem{zhang2019defensepointnet}
Yu Zhang, Gongbo Liang, Tawfiq Salem, and Nathan Jacobs.
\newblock Defense-pointnet: Protecting pointnet against adversarial attacks.
\newblock In {\em {IEEE} International Conference on Big Data}, 2019.

\bibitem{zhao2020isometry}
Yue Zhao, Yuwei Wu, Caihua Chen, and Andrew Lim.
\newblock On isometry robustness of deep 3d point cloud models under
  adversarial attacks.
\newblock In {\em {IEEE/CVF} Conference on Computer Vision and Pattern
  Recognition}, 2020.

\bibitem{zhou2019dupnet}
Hang Zhou, Kejiang Chen, Weiming Zhang, Han Fang, Wenbo Zhou, and Nenghai Yu.
\newblock Dup-net: Denoiser and upsampler network for 3d adversarial point
  clouds defense.
\newblock In {\em {IEEE/CVF} International Conference on Computer Vision},
  2019.

\end{thebibliography}
